\relax
\documentclass[letterpaper]{article} 
\usepackage{aaai22}  
\usepackage{times}  
\usepackage{helvet}  
\usepackage{courier}  
\usepackage[hyphens]{url}  
\usepackage{graphicx} 
\usepackage{natbib}  
\usepackage{caption} 
\DeclareCaptionStyle{ruled}{labelfont=normalfont,labelsep=colon,strut=off} 
\usepackage{algorithm}
\usepackage{algorithmic}

\usepackage{hyperref}       
\usepackage{url}            
\usepackage{booktabs}       
\usepackage{amsfonts}       
\usepackage{nicefrac}       
\usepackage{microtype}      
\usepackage{xcolor}         
\usepackage{amsmath}
\usepackage{wrapfig}
\usepackage{subcaption}
\usepackage{amsthm}

\newtheorem{lemma}{Lemma}
\newtheorem{definition}{Definition}[section]
\usepackage{multirow}
\usepackage{paralist}

\usepackage{pifont}
\newcommand{\cmark}{\ding{51}}%
\newcommand{\xmark}{\ding{55}}%
\usepackage{newfloat}
\usepackage{listings}
\floatstyle{ruled}
\newfloat{listing}{tb}{lst}{}
\floatname{listing}{Listing}

\title{Separating Boundary Points via Structural Regularization for Very Compact Clusters}

\author {
    Xin Ma,\textsuperscript{\rm 1}
    Won Hwa Kim \textsuperscript{\rm 1,2}
}
\affiliations {
    \textsuperscript{\rm 1} University of Texas at Arlington, USA\\
    \textsuperscript{\rm 2} POSTECH, South Korea\\
    \texttt{xin.ma@mavs.uta.edu}, \texttt{wonhwa@postech.ac.kr}
}

\usepackage{bibentry}

\begin{document}
\maketitle

\begin{abstract}
Clustering algorithms have significantly improved along with Deep Neural Networks which provide effective representation of data. Existing methods are built upon deep autoencoder and self-training process that leverages the distribution of cluster assignments of samples. However, as the fundamental objective of the autoencoder is focused on efficient data reconstruction, the learnt space may be sub-optimal for clustering. Moreover, it requires highly effective codes (i.e., representation) of data, otherwise the initial cluster centers often cause stability issues during self-training. Many state-of-the-art clustering algorithms use convolution operation to extract efficient codes but their applications are limited to image data. In this regard, we propose an end-to-end deep clustering algorithm, i.e., Very Compact Clusters (VCC). VCC takes advantage of distributions of local relationships of samples near the boundary of clusters, so that they can be properly separated and pulled to cluster centers to form compact clusters. Experimental results on various datasets illustrate that our proposed approach achieves competitive clustering performance against most of the state-of-the-art clustering methods for both image and non-image data, and its results can be easily qualitatively seen in the learnt low-dimensional space. 
\end{abstract}
 
\vspace{-5pt}
\section{Introduction}
\label{sec:introduction}

Clustering 
aims to separate scattered $N$ data samples $\mathcal{X}=\{\mathbf X_{i}\}_{i=1}^{N}$ in a feature space  
into different groups (e.g., $K$ number of clusters) in an unsupervised way. 
In general, the priority is to 
gather the samples within the same group  
close and make the samples across different clusters distinct from each other. 
As a fundamental topic in machine learning, 
clustering has played a critical 
role 
in a broad range of fields 
including gene sequence clustering in bioinformatics~\cite{petegrosso2020machine,zou2020sequence}, creation of perfectionism profiles in social science~\cite{bolin2014applications}, unsupervised image segmentation~\cite{kanezaki2018unsupervised,ji2019invariant}, document clustering in information retrieval~\cite{xu2003document,fard2020seed,costa2020document} and etc.

Traditional intuitive methods such as $k$-means~\cite{kmeans}, DenPeak~\cite{denpeak,yaohui2017adaptive}, DBSCAN~\cite{dbscan}, and Spectral Clustering~\cite{stsc,scls} 
have been effective in the past decades when datasets used to be relatively small. 
As recent datasets become much larger both in their size and dimension, 
the traditional shallow methods suffer from 
high computational complexity and decrease in performance.
Various Deep Clustering (DC) techniques have been recently developed to cope with issues with conventional approaches. 
The core of DC methods consists of two components: 
1) Dimension Reduction with Deep Learning (e.g., autoencoder) for mapping high-dimensional data 
onto a low-dimensional space 
and 2) Self-training the low-dimensional embedding to further improve clustering results using traditional clustering algorithms such as $k$-means \cite{xie2016unsupervised,ghasedi2017deep}.   

The premise behind the DC algorithms is that a
suitable low-dimensional embedding and cluster centers 
will assign each sample to a vivid individual cluster. 
These techniques focus on optimizing `cluster assignment probability', 
which is a likelihood of a sample belong to each cluster. 
This measure is often defined by the distance between 
the sample and cluster centers; 
DC methods minimize Kullback–Leibler (KL) divergence between 
the distribution of the cluster assignment probability and an auxiliary target distribution directly computed from it.  
Such a process makes the assignment probability localized to a single cluster for each sample.

\begin{figure*}[!t]
\begin{minipage}[!t]{1.0\textwidth}
  \centering
  \hspace{-13mm}
    \begin{subfigure}{.3\textwidth}
        \centering
        \includegraphics[height = 2.3cm]{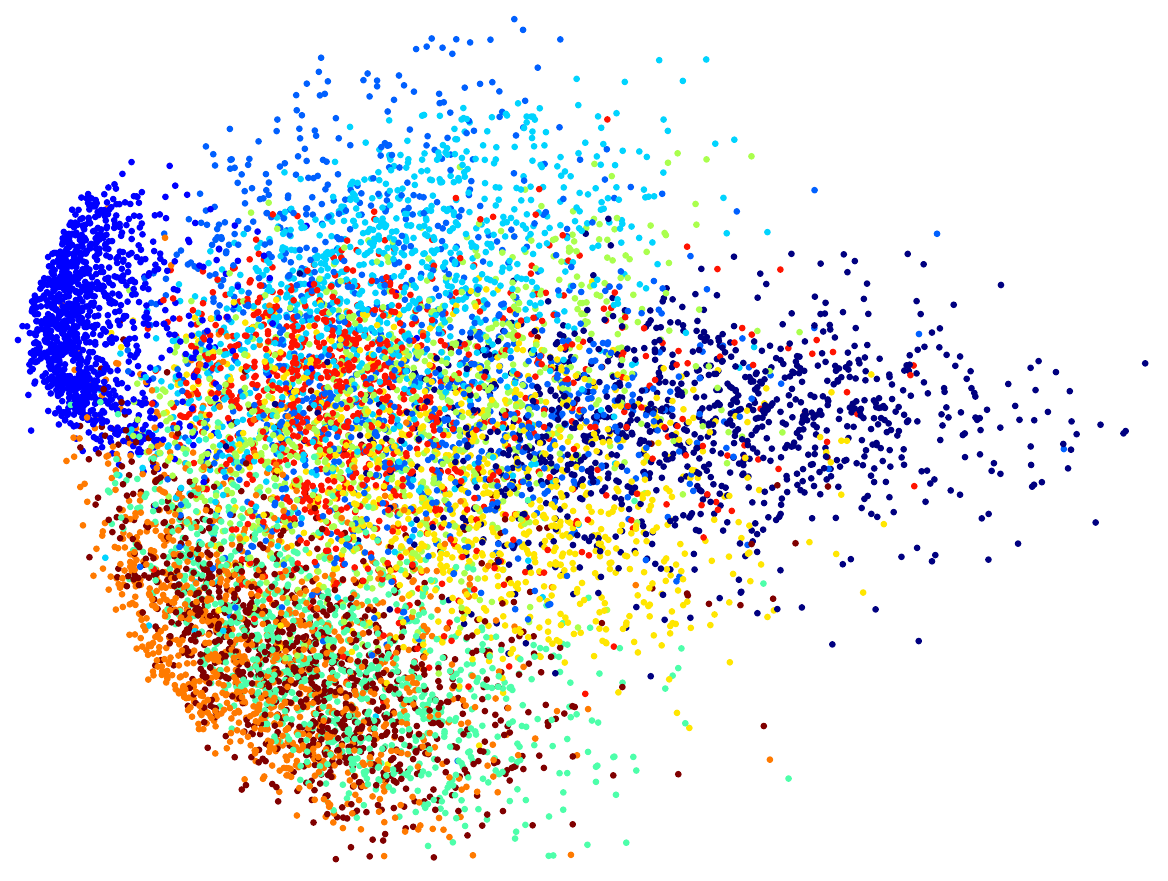}
        \caption{\footnotesize Raw Data.}
    \end{subfigure}\hspace{-8mm}
    \begin{subfigure}{.3\textwidth}
        \centering
        \includegraphics[height = 2.3cm]{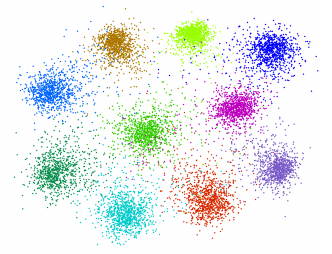}
        \caption{\footnotesize JULE.}
    \end{subfigure}\hspace{-8mm}
    \begin{subfigure}{.3\textwidth}
        \centering
        \includegraphics[height = 2.3cm]{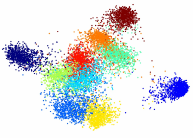}
        \caption{\footnotesize DEPICT.}
    \end{subfigure}\hspace{-8mm}
    \begin{subfigure}{.3\textwidth}
        \centering
        \includegraphics[height = 2.3cm]{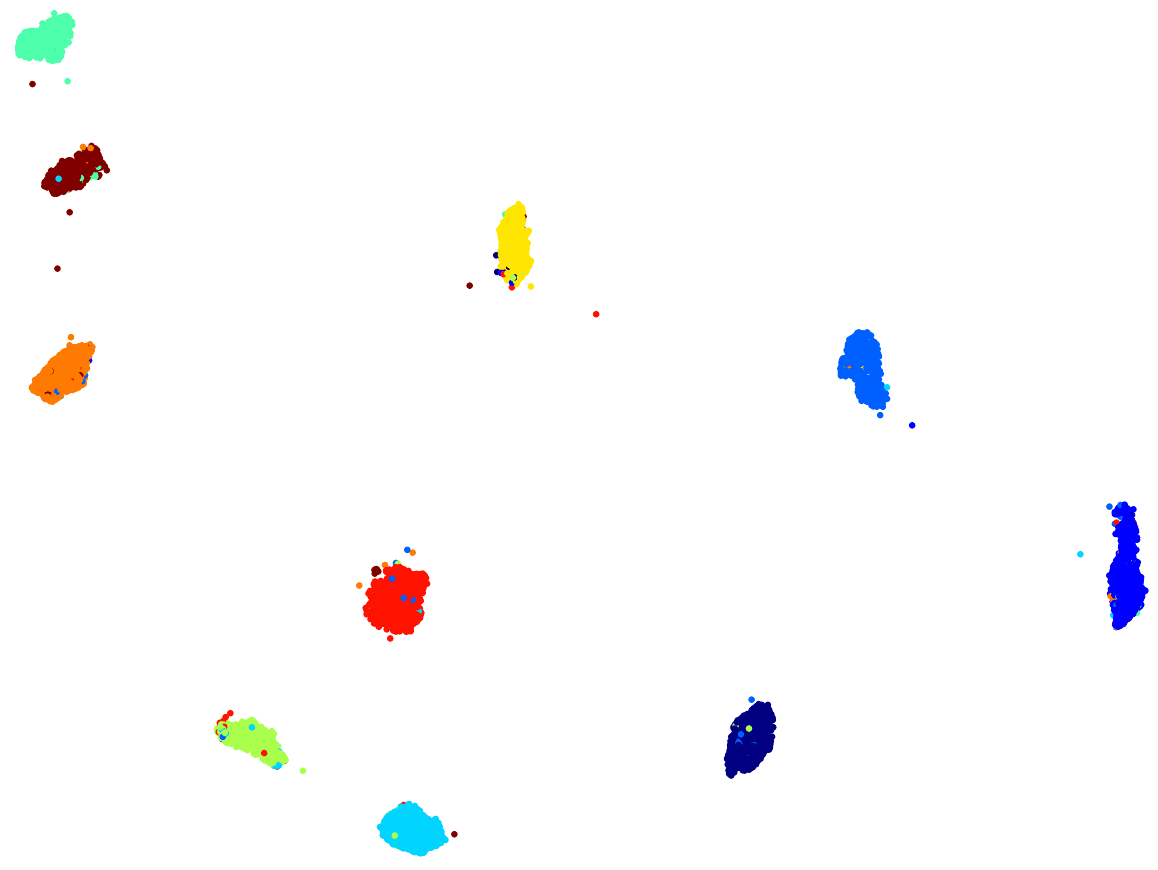}
        \caption{\footnotesize VCC (Ours).}
    \end{subfigure}\hspace{-10mm}
    \end{minipage}
    \caption{\footnotesize Visualization of embedding in subspaces optimized by different methods on MNIST-test dataset. (a) The raw data in 2D with PCA. (b) The embedding subspace of JULE~\cite{jule}. (c) The embedding subspace of joint DEPICT~\cite{ghasedi2017deep}. (d) The embedding subspace of VCC by well handling with boundary points using graph structure regularization.
    }
    \vspace{-15pt}
    \label{fig:synthetic_demo}
\end{figure*}

Unfortunately, many DC approaches suffer from two major bottlenecks. 
First, the initial cluster centers provided by a centroid-based clustering algorithm with {\em random} initialization, e.g., $k$-means, often cause stability issues, especially when the low-dimensional embedding is not sufficiently effective for clustering. 
For imaging data,
convolutions are used 
to learn a better feature space for the initial clustering \cite{ghasedi2017deep,ren2020deep}, but their usages are limited to images. 
The second issue, perhaps even more critical, comes from their self-training process 
where the update of cluster assignment probability 
mainly focuses on ``easy samples'' close to the centers and overlooks the samples near the {\em boundary} of clusters. 
Such approaches require a clustering-oriented low-dimensional embedding, 
which autoencoder may not provide, as the autoencoder mainly focuses on preserving variations in the data.

In fact, for clustering, 
what matters over the variation in the data is 
the relationships between data points (i.e., structure) typically represented as a graph, e.g., $k$-nearest neighbor ($k$NN) graph.
Therefore, an ideal low-dimensional space should preserve the inherent graph structure, which an autoencoder may overlook. 
We also hypothesize that the actual similarity measures between data points 
will make a downstream clustering result sub-optimal. 
Notice that an ideal cluster would have all the samples in the same cluster {\em compactly} merged at the center, but optimization based on the similarity measures prevent those samples at the {\em boundary of clusters} from being pulled to the cluster center.
For the compact clustering, we propose the following ideas: 
1) the algorithm should focus more on the relatively sparse samples at the cluster boundary 
with {\em skewed} distribution of similarities 
as opposed to the densely populated samples near cluster centers with similarities with {\em less variation}, 
and 2) only the connection information between the samples should be considered to ultimately attract samples to cluster centers. 

For this, we propose an end-to-end clustering algorithm that implements our ideas above. 
Our framework specifically focuses more on samples at the boundaries of clusters via {\em sampling} during optimization and achieves a desirable latent space
for clustering using {\em structural regularizers}. 
Our work demonstrates the following {\bf contributions:} 
\begin{compactenum}[\bfseries (i)]
\item we propose a novel method ``Very Compact Clusters'' (VCC) that 
achieve {\em very compact clusters} by operating on samples near cluster boundaries, 
\item VCC performs dimension reduction, self-training and clustering simultaneously as a unified framework, 
\item we carry extensive empirical validation of VCC with various independent datasets, which demonstrate 
competitive qualitative and quantitative performances. 
\end{compactenum}
Fig.~\ref{fig:synthetic_demo} (d) is a teasing result from VCC achieving localized clusters in a learnt 2D
space using high-dimensional MNIST-test data, and its details are introduced in the following. 

\vspace{-5pt}
\section{Related Work}

\label{sec:related_work}

Deep Clustering introduces deep learning 
into clustering to learn effective representations and cluster assignments ~\cite{guo2019adaptive,xie2016unsupervised,yang2019deep,huang2020deep}. 
Xie et al. proposed deep embedded clustering (DEC) to jointly perform low-dimensional embedding optimization and update cluster centers~\cite{xie2016unsupervised}. Dizaji et al. utilized a denoised autoencoder to further improve the low-dimensional embedding and achieved excellent clustering performance~\cite{ghasedi2017deep}. Chang et al. introduced the convolution neural network to deep clustering and achieved high clustering performance on image datasets~\cite{chang2017deep}. Ren et al. proposed to use a density-based clustering algorithm (e.g., DenPeak) to initialize cluster centers and obtained good performance on image cluster discovery~\cite{ren2020deep}. Some of these methods demonstrated even more powerful results together with data augmentation. SCAN~\cite{van2020scan} and MiCE~\cite{tsai2020mice} recently achieved state-of-the-art clustering performance on image datasets using contrastive learning. 

{\em Benefits from our work.} 
We take advantage of the relationships among samples to identify boundary points between clusters. 
Our framework 
takes care of boundary points by leveraging distribution of similarities 
and perform clustering to generate highly compact 
clusters in the latent space. 

\vspace{-5pt}
\section{VCC: Very Compact Clusters}
\label{sec:methods}

Given a dataset $\mathcal{X} = \{ \mathbf{X}_{i} \in \mathbb{R}^{D}\}_{i}^{N}$ with $N$ samples in $D$-dimensional space, the 
principle of clustering is to separate $\mathcal{X}$ into $K$ clusters such that
intra-cluster samples 
stay compact while inter-cluster samples stay far apart. 
VCC aims to obtain compact clustering in a low-dimensional latent space by emphasizing samples at cluster boundaries. 

\vspace{-2pt}
\subsection{Boundary Points Separation with Local Similarity}

Most DC approaches rely on deep autoencoder to find an initial low-dimensional space  \cite{xie2016unsupervised,ren2020deep}. 
However, this choice of the initial latent space is {\em sub-optimal} for clustering as the trained space is mainly focused on 
efficiently representing the original data. 
Intuitively speaking, to achieve good clustering, 
the clusters in the represented space should be distinct from each other with separated cluster boundaries. 
For this, the emphasis of the 
latent space should focus on the {\em samples at the boundary} of clusters rather than highlighting reconstruction of the data.

To illustrate how boundary points affect clustering, let us define boundary points as
\begin{definition}{(Boundary Points Between Clusters)}
\label{definition1}
Let $\mathcal{C}_{A}$ and $\mathcal{C}_{B}$ be two clusters and $\mathcal{N}_{i}$ represent some neighborhood of a sample $\mathbf{X}_{i}$ in data $\mathcal{X}$. The $\mathbf{X}_{i}$ is a boundary point if it satisfies the following conditions:
\begin{enumerate}
    \item $\mathbf{X}_{i}$ is in a densely populated region $\mathbb{R}$ in $\mathcal{X}$;
    \item $\exists$ a region with much lower or higher population of samples $\mathbb{R}^{\prime}$ near $\mathbf{X}_{i}$; 
    \item $\exists$ $\mathbf{X}_{j}$ and $\mathbf{X}_{k}$, $j\neq k, \mathbf{X}_{j}, \mathbf{X}_{k} \in \mathcal{N}_{i}$ such that $\mathbf{X}_{j} \in \mathcal{C}_{A}~~ \text{and}~~ \mathbf{X}_{k} \in \mathcal{C}_{B}$ 
\end{enumerate}
\end{definition}%

Using the Definition \ref{definition1}, the Lemma \ref{lem:lem1} (proof given in the supplementary) 
tells that we can distinguish interior points (i.e., samples) near cluster centers and boundary points by leveraging the {\em variance difference of the local structure of data samples} in the dataset. 
This is a key observation as separating samples near the cluster boundaries is critical in obtaining accurate clustering. 

\begin{lemma}
\label{lem:lem1}

Let $\mathcal{X}_{I}$ and $\mathcal{X}_{J}$ be bounded separable subsets of a dataset $\mathcal{X}$ in its feature space. Assume $\mathbf{X}_{i}$ and $\mathbf{X}_{j}$ are two interior points of $\mathcal{X}_{I}$ and $\mathcal{X}_{J}$, and $\mathbf{X}_{t}$ is a boundary point between $\mathcal{X}_{I}$ and $\mathcal{X}_{J}$ defined in Definition~\ref{definition1} with a region $\mathcal{X}_{R}=\{\mathbf{X}_{l}~|~ d(\mathbf{X}_{t}, \mathbf{X}_{l}) < \rho, \rho > 0\}$ where $d(\cdot)$ is a distance metric. 
Let $\mathcal{D}_{I}$, $\mathcal{D}_{J}$ and $\mathcal{D}_{T}$ be sets of $M$ nearest distances around $\mathbf{X}_{i}$, $\mathbf{X}_{j}$ and $\mathbf{X}_{t}$, 
Then, $Var(\mathcal{D}_{T}) > Var(\mathcal{D}_{I})$ or $Var(\mathcal{D}_{T}) > Var(\mathcal{D}_{J})$.
\end{lemma}

\begin{figure}[!tb]
\vspace{-5pt}
  \centering
    \includegraphics[width=.8\linewidth]{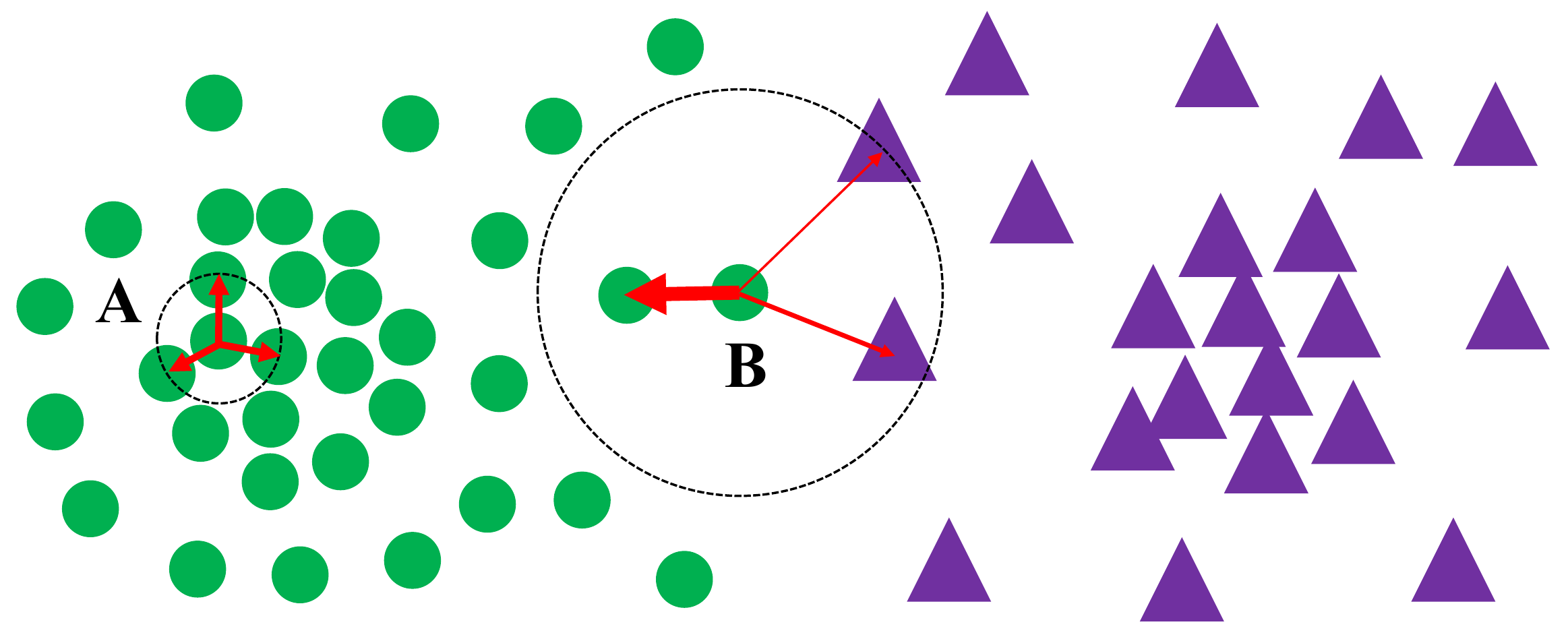}
    \caption{\footnotesize Illustration of edge weights distribution difference between boundary point (i.e., B) and interior point (i.e., A) in a $k$-NN graph ($k=3$). Line width represents the edge weight.}
    \vspace{-10pt}
    \label{fig:illustration}
\end{figure}

Separating boundary points is critical for many clustering tasks (e.g., centroid-based methods). Lemma \ref{lem:lem1} shows a way to identify boundary points in the dataset with the variance of $M$ nearest neighbors' distances of each point. 
That is, a point with large variance in the distances among its nearest neighbors is likely to be a boundary point.
Considering that samples near a boundary point are sparser than the samples near a center (see Fig.~\ref{fig:illustration}), 
normalization is required to make sure that 
similarities in the local neighborhood are determined by relative distances instead of globally absolute distances.

\vspace{-2pt}
\paragraph{Structure via Local Similarity in High-dimensional Space.} 
A $k$NN graph, which connects $k$ nearest neighbors of individual samples, is often used to capture local relationships among data points and reveal global structure in a dataset \cite{ding2004k}. However, a naively constructed $k$NN graph focuses only on the most similar samples while ignores the boundary samples when clustering. Inspired by UMAP \cite{mcinnes2018umap} which also uses graph structure for embedding, here, we construct a normalized latent graph (LG) $\mathcal G_{lg}$, whose vertices are data samples and edge weights are defined in the following. 

Consider a distance matrix $\mathbf{D}_{N\times M}$ computed from $\mathcal{X}$.
The elements in each row $\mathbf{D}_{i:}, i\in [1, 2, ..., N]$ of $\mathbf{D}$ are distance metrics (e.g., Euclidean distance) of the $i$-th sample to its $M$ different nearest neighbors (we use $M$ instead of $k$ to avoid confusion with the number of clusters $K$). 
As discussed above, the edge weights need to be normalized and 
the variations of edge weights around the boundary point should become large. 
Therefore, we use
a softmax function at each row of $\mathbf{D}$
to yield a locally normalized distance vector, $\mathbf{F}_{i}$, whose elements quantify the ``attractive forces'' at the $i$-th individual node:
\begin{equation}
\small
    \label{eq:softmax_force}
    \mathbf{F}_{i:} =  \text{softmax}(- \mathbf{D}_{i:}).
\end{equation}

Note that $\mathbf{F}_{ij}$ can also be regarded as the probability that $\mathbf{X}_i$ and $\mathbf{X}_j$ are connected with the direction from $i$ to $j$.
From $\mathbf{F}_{ij}$, an undirected and symmetric adjacency matrix $\Tilde{\mathbf{F}}$ is derived as
\begin{eqnarray}
\small
    \label{eq:knn}
    \Tilde{\mathbf{F}}_{ij} & = &  1 - (1-\mathbf{F}_{ij})(1-\mathbf{F}_{ji}) \nonumber\\
    & = & \mathbf{F}_{ij} + \mathbf{F}_{ji} - \mathbf{F}_{ij}\mathbf{F}_{ji}.
\end{eqnarray}
which become edge weights of a latent graph $\mathcal{G}_{lg}$. 
Making $\Tilde{\mathbf{F}}$ imputes edges for the samples whose relationships were one-sided in $\mathbf{F}$, 
and makes separation of boundary points effective by introducing connections to other interior samples.

\vspace{-2pt}
\paragraph{Similarities in Low-dimensional Space.} 
Many DC techniques project the original high-dimensional samples onto a low-dimensional latent space for efficiency. 
Let $\mathcal{H}$ be an unknown embedding of samples in a low-dimensional latent space. 
It will be beneficial for clustering if separation of the boundary points can be incorporated when learning the $\mathcal{H}$. 
Separating boundary points in $\mathcal{H}$ can be done by optimizing the hidden structure (e.g., similarity) of $\mathcal{H}$. 
To define this hidden structure, we first need a metric to quantify similarities among samples in $\mathcal{H}$.

Given two samples $\mathbf{H}_{i}$ and $\mathbf{H}_{j}$ from $\mathcal{H}$,  
their similarity $\mathbf{\mathring F}_{ij}$ should be antidependent on their distance $d(\mathbf{H}_{i}, \mathbf{H}_{j})$, 
and we use an exponential function to define $\mathbf{\mathring F}$ as 
\begin{equation}
\small
    \label{eq:smooth_approximation}
    	\mathbf{\mathring F}_{ij} = e^{-d(\mathbf{H}_{i}, \mathbf{H}_{j})}. 
\end{equation}
Here, two close samples (i.e., with small $d(\mathbf{H}_{i}, \mathbf{H}_{j})$) leads to a large $\mathbf{\mathring F}_{ij}$ which indicates the attractive force between those two samples is strong in the latent space, and vice versa for two far samples. 

\vspace{-2pt}
\paragraph{Boundary Points Separation Loss.}
Given a latent graph constructed by 
\eqref{eq:knn}, one can see that small edge weights in the latent graph are closely related to boundary points between different clusters. 
The reason is seen in \eqref{eq:softmax_force} 
that normalizes the edge weights of $k$NN sub-graph at each point. 
At the central region of each cluster, 
samples stay close to each other and 
the weights in their sub-graphs have low variation.
On the contrary, at the cluster boundaries, 
samples are rather sparsely distributed 
and it is highly likely to obtain imbalanced sub-graphs  
with large average and variations in their distances. 
Fig.~\ref{fig:illustration} illustrates such a behavior, 
where samples around a boundary point $\bf B$ are sparse with various similarities (edge thickness)  
while samples around an interior point  $\bf A$ are populated with low variation in their similarities. 

However, operating on all edges in a graph is challenging since the number of edges increases dramatically along with dataset size.
To reduce the computation, a sampling-based loss is adopted here. 
Given the observation that the boundary points are highly associated with small edge weights, 
we define a sampling rate $\mathbf{R}_{ij}^{-} \in \mathbb{Z}^{0+}$ for epoch (i.e., the number of samples per epoch) at 
the edge $(\mathbf{X}_{i}, \mathbf{X}_{j})$ 
as  
\begin{equation}
\small
    \label{eq:sampling}
    \mathbf{R}_{ij}^{-} = 	\left\lfloor\left(\frac{\Tilde{\mathbf{F}}_{max}}{\Tilde{\mathbf{F}}_{ij}}\right)\right\rfloor
\end{equation}
where $\lfloor \cdot \rfloor$ is the floor function and $\Tilde{\mathbf{F}}_{max}$ is the largest value in $\Tilde{\mathbf{F}}$.  
The $\mathbf{R}^{-}$ lets the edges with small weights become more likely to be sampled to 
compute $\mathcal{L}_{bps}$. Here, we utilize the batch-wise training where the sampling rate $\mathbf{R}^{-}$ is guaranteed in the form of the ratio of the number of edges in each batch. To achieve this, we create an augmented edge set $\mathcal{E}^{-}$ by duplicating each edge indicated in $\mathcal{G}_{lg}$ according to $\mathbf{R}_{ij}^{-}$. 
Finally, boundary points can be separated by minimizing
\begin{equation}
\small
    \label{eq:bp_loss_sampling}
    \mathcal{L}_{bps} = - \underset{(i,j)\in \mathcal{E}^{-}}{\sum} \log(\mathbf{\mathring F}_{ij}) +  \log(1 - \mathbf{\mathring F}_{ij}).
\end{equation}
Here, for each pair of samples ($\mathbf{X}_{i}$, $\mathbf{X}_{j}$), $\log(\mathbf{\mathring F}_{ij})$ attracts while $\log(1 - \mathbf{\mathring F}_{ij})$ expands them. Moreover, the imbalance of edges augmented by negative sampling 
with $\mathbf{R}_{ij}^{-}$ forces the model (e.g., MLP) to focus on $\log(1 - \mathbf{\mathring F}_{ij})$ rather than $\log(\mathbf{\mathring F}_{ij})$ in each batch. 
The boundary points are separated 
by strong expansion and weak attraction among their neighbors.

\vspace{-2pt}
\subsection{Learning Embeddings for Compact Clustering}

Minimizing $\mathcal{L}_{bps}$ yields a low-dimensional latent space where boundary points of clusters can be effectively separated. 
It preserves large variation 
in similarity 
within the same cluster (i.e., local structure) 
as well as 
between different clusters (i.e., global structure). 
While preserving the structure in the data can be useful, 
an accurate clustering may be difficult 
for those points with low similarities 
among samples in the same cluster but high similarities with samples in other clusters. 

Here, our solution is to further neglect the local structure. 
This is because the shape of an ideal cluster would be a single point, 
i.e., all samples merged at the center of each cluster, but preserving local similarity will definitely impede such behavior. 
Based on this idea, 
we develop a framework called ``\textit{\textbf{Very Compact Clusters (VCC)}}'' (See Fig.~\ref{fig:synthetic_demo} (d)). 
To achieve compact clusters, two more 
concepts are introduced: \textit{Contraction} and \textit{Expansion}. 
The contraction makes intra-cluster samples compactly gathered, while the expansion pushes inter-cluster samples apart. 
Note that VCC is a principle used for clustering, and the compactness of the final clustering result still depends on the quality of input data. 

\vspace{-2pt}
\paragraph{Contraction Loss.} 
To attract samples in each cluster to their centers, 
we emphasize the edges with large weights which indicate strong attractive forces to attract neighboring samples together. 
Again, due to high computation from getting exhaustive pair-wise distances, we opt to perform a sampling-based optimization. 
The sampling strategy is based on the sampling rate $\mathbf{R}_{ij}^{+} \in \mathbb{Z}^{0+}$ on epoch at each edge between the $i$-th and $j$-th nodes defined as
\begin{equation}
\small
    \label{eq:positive_sampling}
    \mathbf{R}_{ij}^{+} = 	\left\lfloor\left(\frac{\Tilde{\mathbf{F}}_{ij}}{\Tilde{\mathbf{F}}_{mean}}\right)\right\rfloor
\end{equation}
where $\Tilde{\mathbf{F}}_{mean}$ is the mean value in $\Tilde{\mathbf{F}}$. From \eqref{eq:positive_sampling}, we can see that a large $\alpha$ leads to a high $\mathbf{R}_{ij}^{+}$, which increases the chances to sample the edges with strong attractive forces during the training. Based on \eqref{eq:positive_sampling}, we create another augmented edge set $\mathcal{E}^{+}$ by duplicating each edge in $\mathcal{G}_{lg}$ according to $\mathbf{R}_{ij}^{+}$.
Then, the contraction loss $\mathcal{L}_{c}$ is given as
{ \begin{eqnarray}
\small
    \label{eq:ec_loss}
    \mathcal{L}_{c} =  - \underset{(i,j)\in \mathcal{E}^{+}}{\sum} \log(\mathbf{\mathring F}_{ij})
\end{eqnarray}}

\vspace{-2pt}
\paragraph{Expansion Loss.} In contrast to the contraction process, to make compact clusters, disconnected samples will be dramatically separated during the expansion process in $\mathcal{H}$. 
Based on 
local similarity from $\Tilde{\mathbf{F}}$, 
the 
local connectivity $\mathbf{B}_{N\times N}$ is computed with a sign function as 
\begin{equation}
\small
    \label{eq:local_connectivity}
    \mathbf{B} = \textnormal{sign}(\Tilde{\mathbf{F}}).
\end{equation}
The $\mathbf{B}$ only tells us which of the samples are connected to each other with binary elements. 
Let $\mathcal{B}$ be the coordinate (COO) representation of $\mathbf{B}$.
Then, the disconnectivity set, $\mathcal{B}_{neg}$, will be the set of edges not in $\mathcal{B}$. 
Let $\mathcal{B}^{*}_{neg}$ is a subset of edges randomly sampled from $\mathcal{B}_{neg}$.
Since all the disconnection weights are equal (i.e., 0), 
the expansion loss $\mathcal{L}_{e}$ can be directly written as 
\begin{eqnarray}
    \label{eq:ee_loss}
    \mathcal{L}_{e} =   - \underset{(i,j)\in \mathcal{B}^{*}_{neg}}{\sum} \log(1 - \mathbf{\mathring F}_{ij}).
\end{eqnarray}

The intuitions of $\mathcal{L}_{bps}$, $\mathcal{L}_{c}$ and $\mathcal{L}_{e}$ 
are summarized below: 
\begin{itemize}
    \item $\mathcal{L}_{bps}$ is inspired by UMAP \cite{mcinnes2018umap}, but simple softmax is used to calculate graph edge weights and sampling based on $\mathbf{R}_{ij}^{-}$ instead of solving complicated optimization as in \cite{mcinnes2018umap} to make graph construction more efficient. 
    \item $\mathcal{L}_{bps}$ alone cannot lead to compact embeddings/clusters. Instead of attracting all connected points, $\mathcal{L}_{c}$  selectively draws points together via skewed sampling with $\mathbf{R}_{ij}^{+}$.
    \item $\mathcal{L}_{e}$ separates disconnected points as far as possible for distinct clusters, as two other losses $\mathcal{L}_{bps}$ and $\mathcal{L}_{c}$ are attracting samples to each other. 
    
\end{itemize}
With the sampling technique mentioned above, $\mathcal{L}_{bps}$, $\mathcal{L}_{c}$ and $\mathcal{L}_{e}$ are 
optimized simultaneously to get compact embeddings.

\begin{figure*}[!t]
\vspace{-10pt}
    \begin{subfigure}{.30\textwidth}
        \centering
        \includegraphics[width=0.9\linewidth]{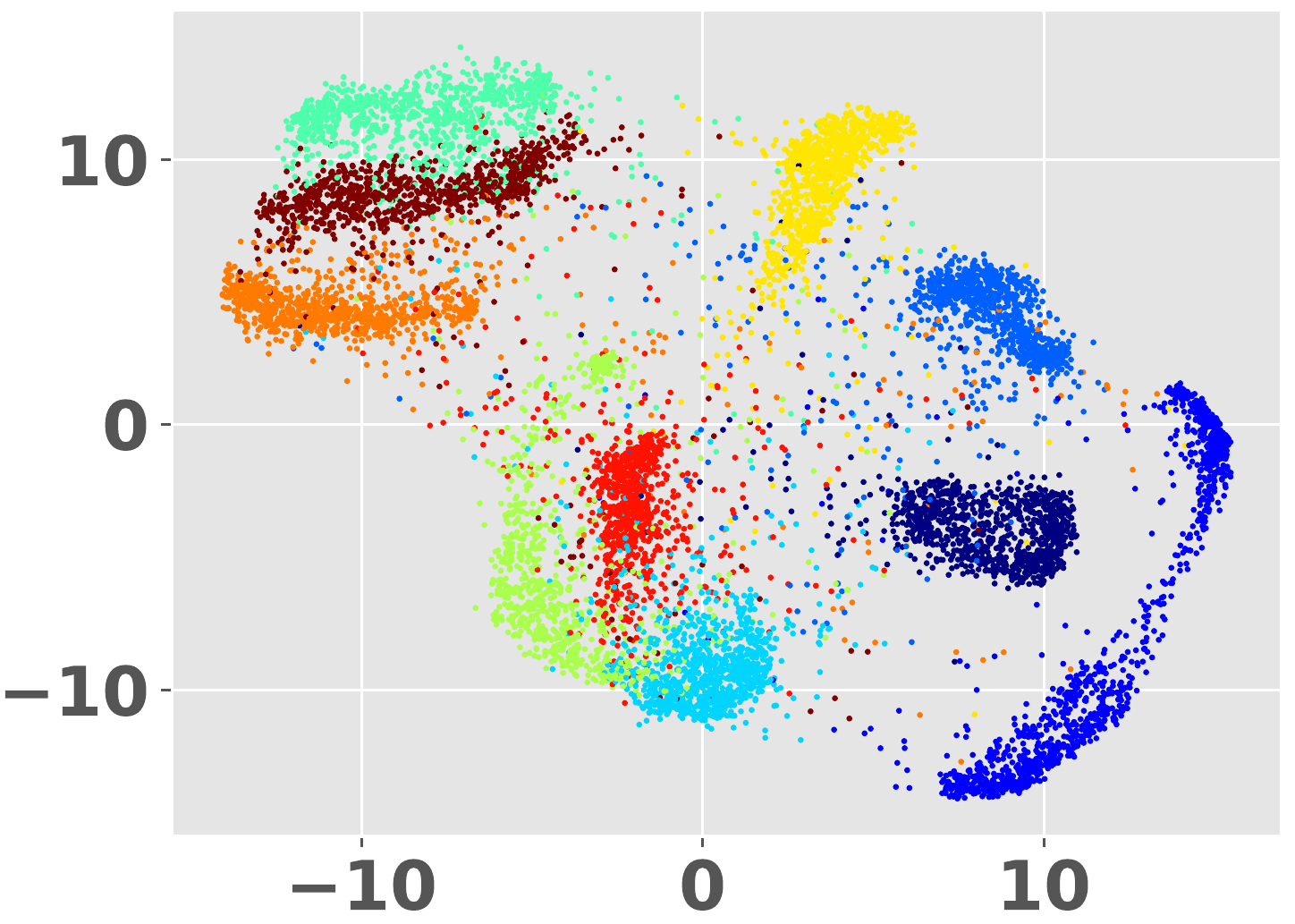}
    \end{subfigure}\hfill
    \begin{subfigure}{.30\textwidth}
        \centering
        \includegraphics[width=0.9\linewidth]{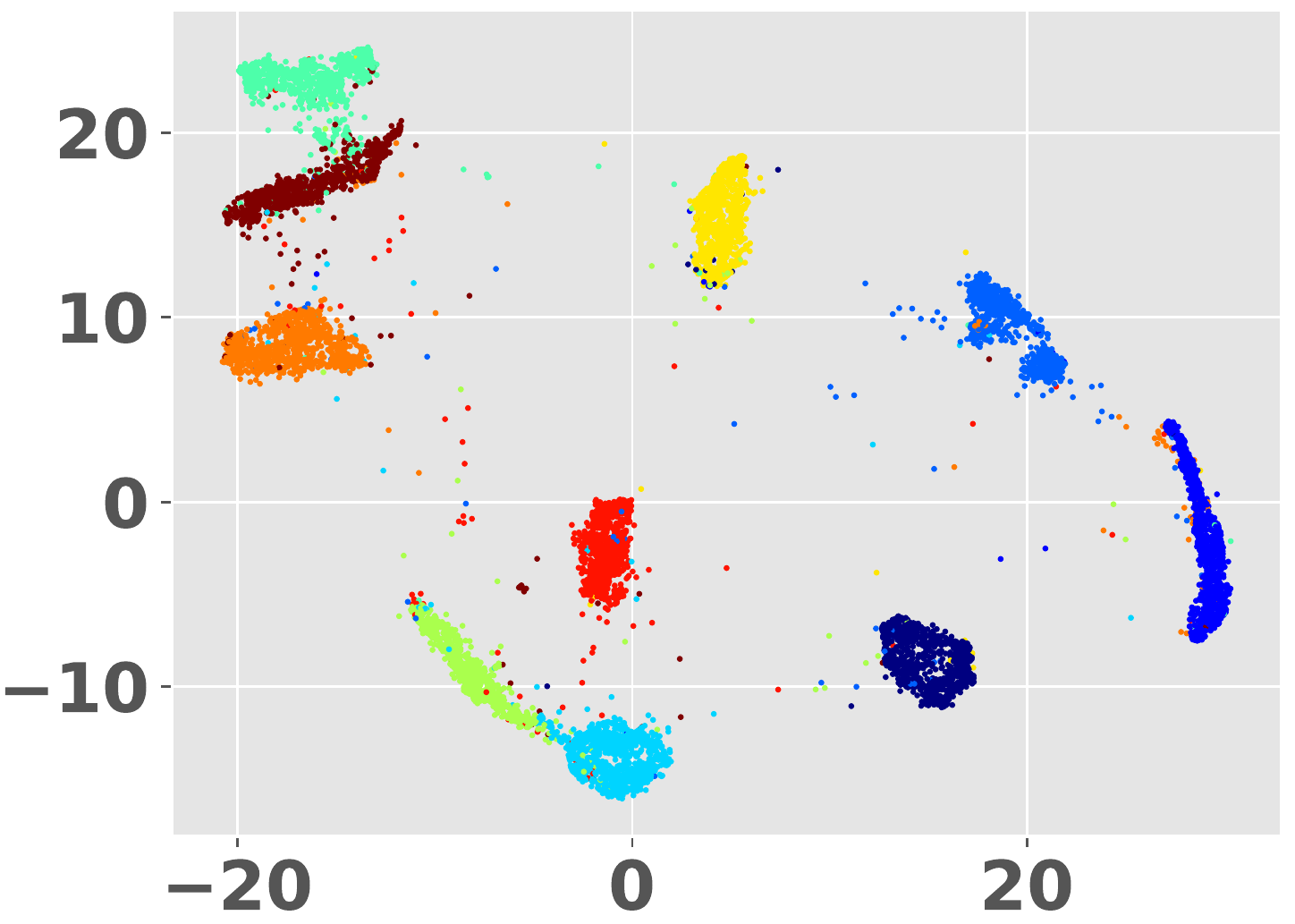}
    \end{subfigure}\hfill
    \begin{subfigure}{.30\textwidth}
        \centering
        \includegraphics[width=0.9\linewidth]{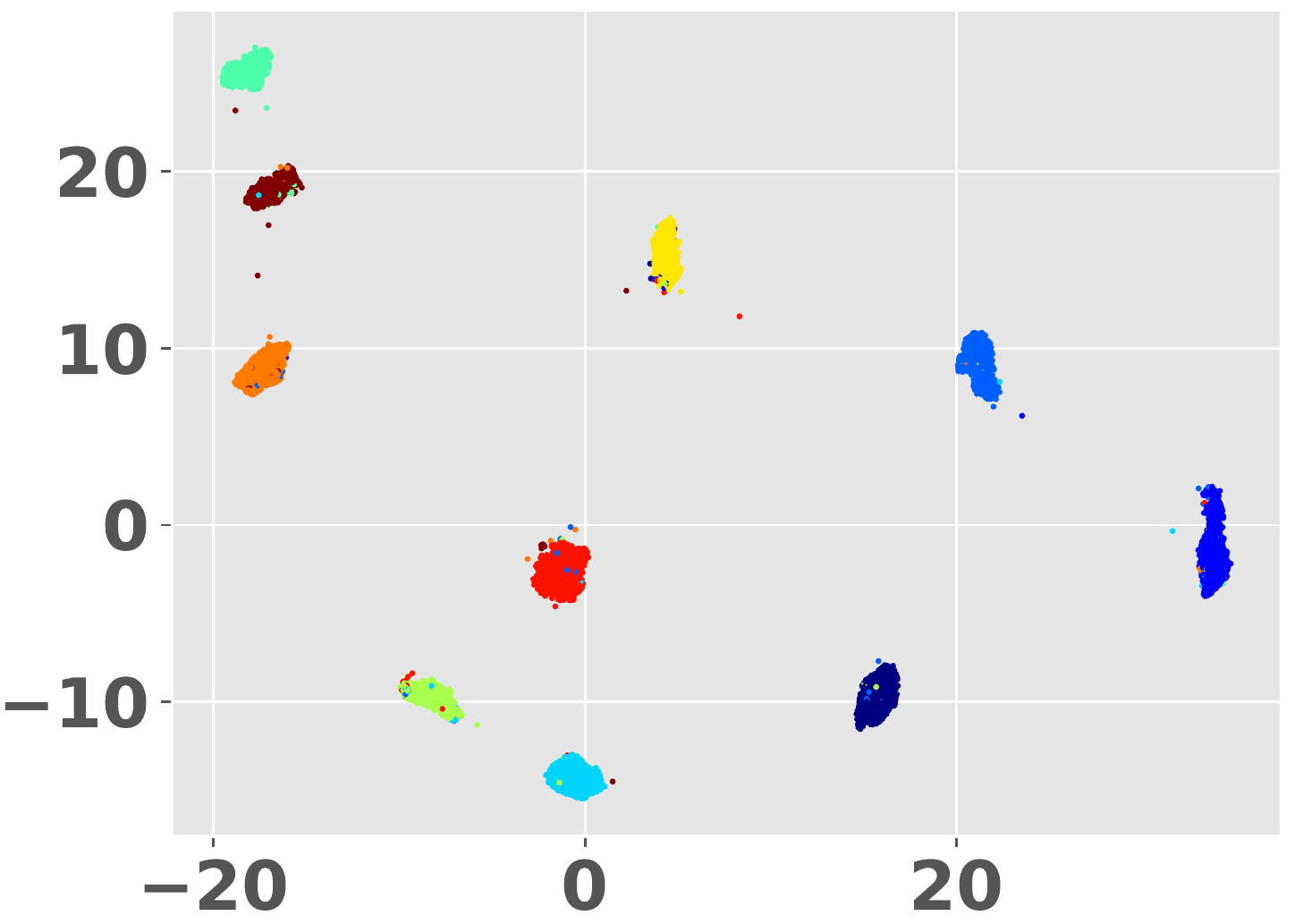}
    \end{subfigure}
    
    \begin{subfigure}{.30\textwidth}
        \centering
        \includegraphics[width=0.98\linewidth]{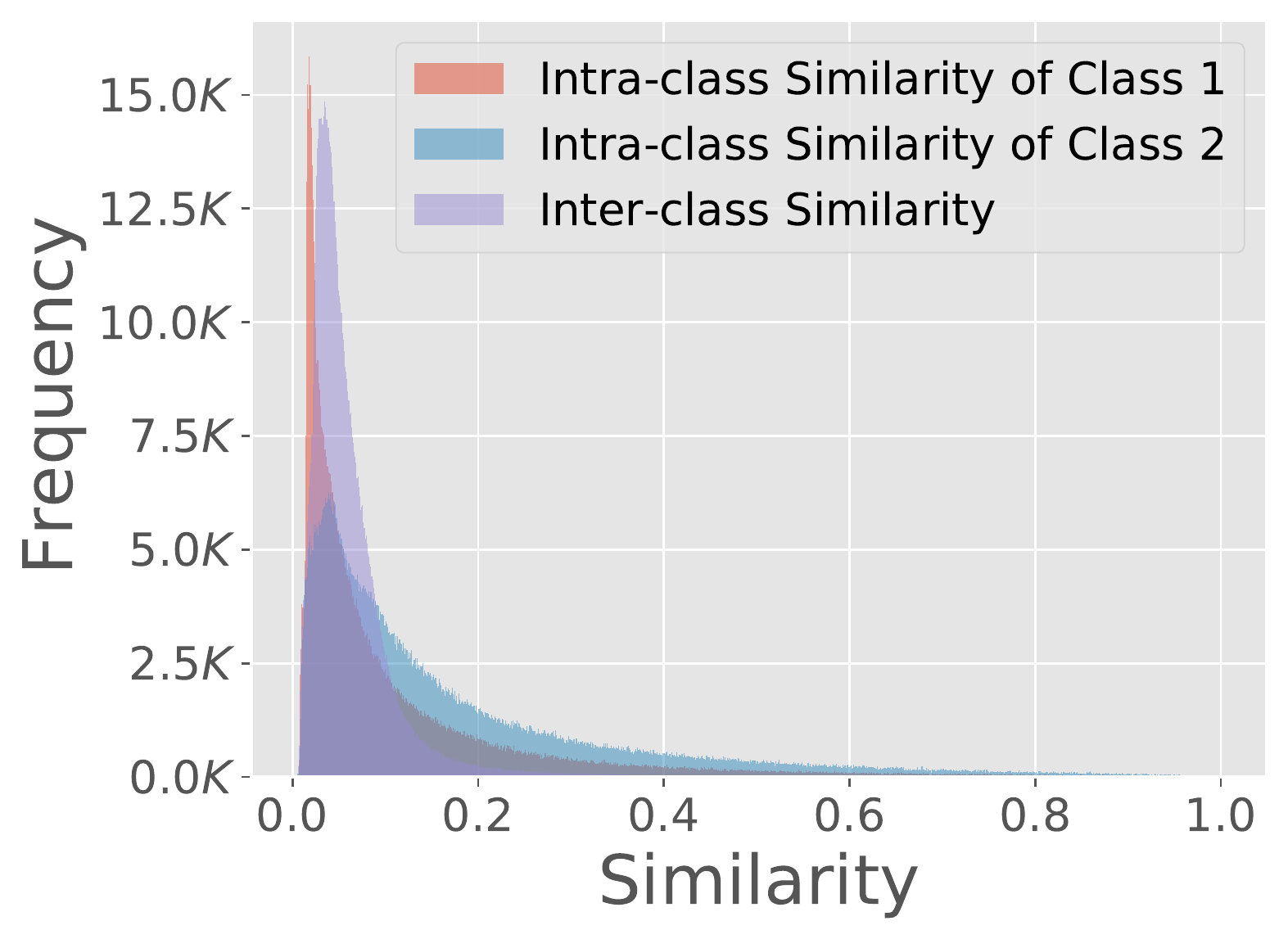}
        \vspace{-5pt}
        \caption{\footnotesize Beginning Epoch.}
    \end{subfigure}\hfill
    \begin{subfigure}{.30\textwidth}
        \centering
        \includegraphics[width=0.98\linewidth]{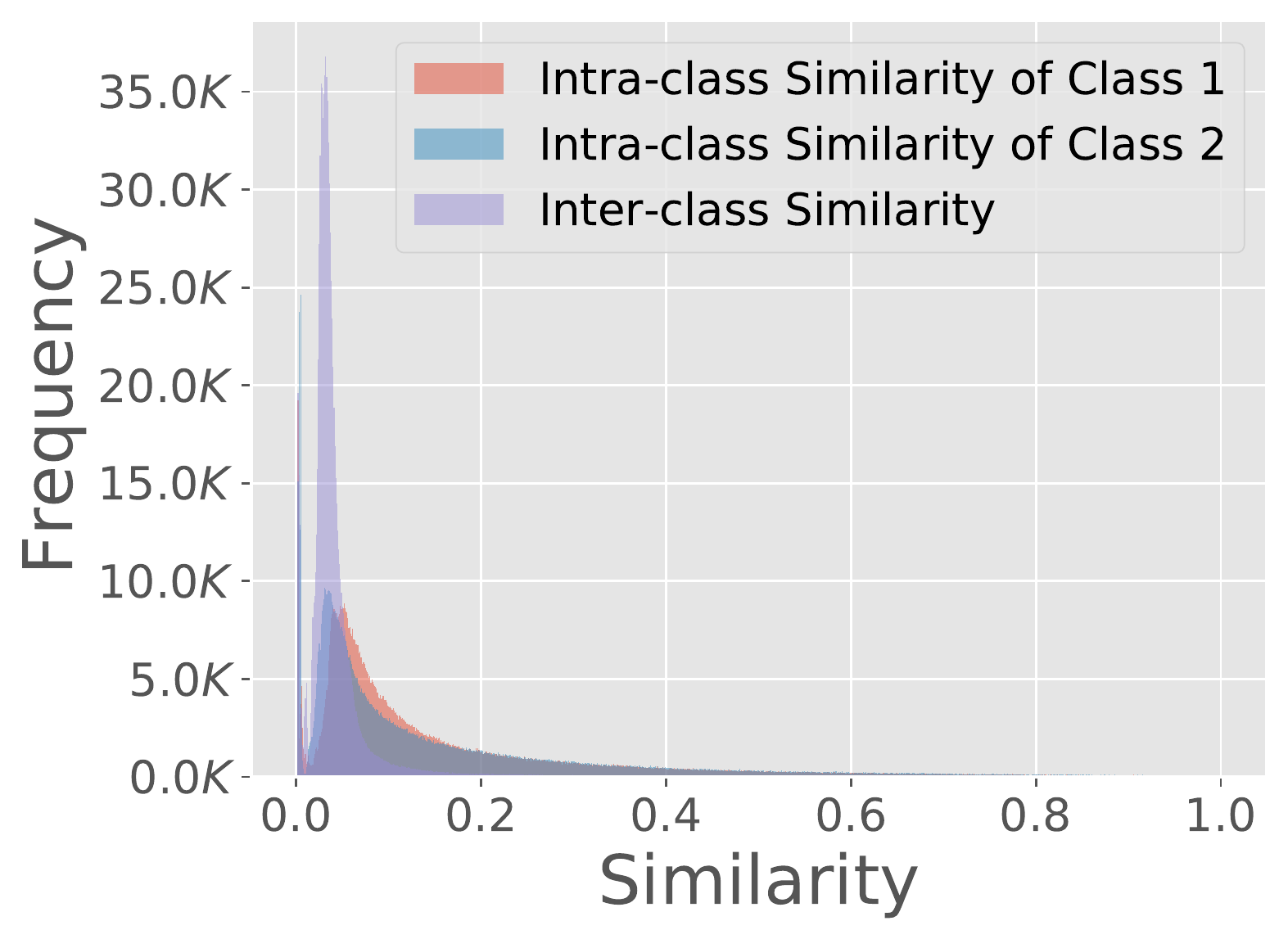}
        \vspace{-5pt}
        \caption{\footnotesize Middle Epoch.}
    \end{subfigure}\hfill
    \begin{subfigure}{.30\textwidth}
        \centering
        \includegraphics[width=0.98\linewidth]{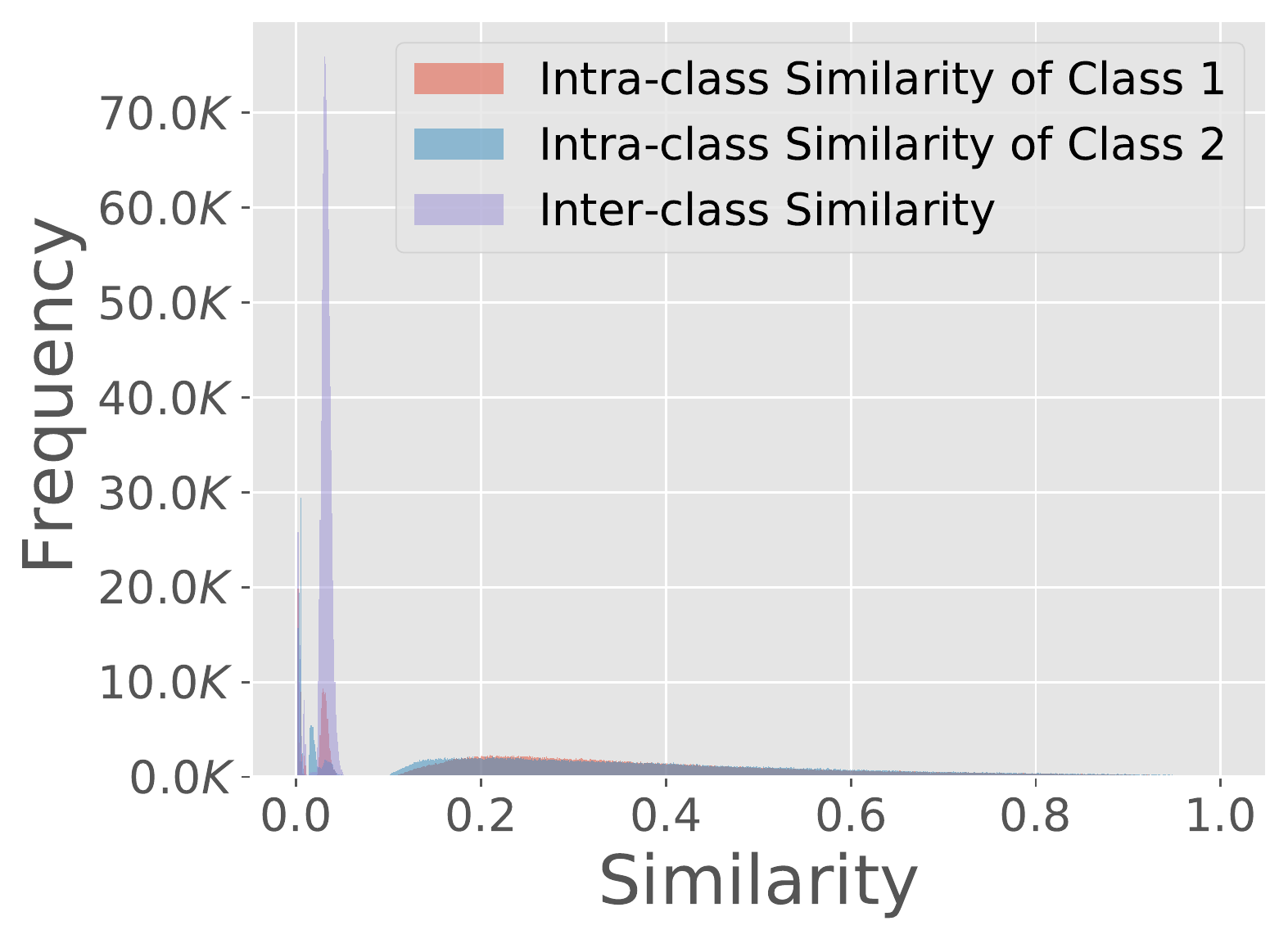}
        \vspace{-5pt}
        \caption{\footnotesize Final Epoch.}
    \end{subfigure}
    \vspace{-8pt}
    \caption{\footnotesize Visualization of VCC's clustering assignment performance on MNIST-test data throughout training (the input embedding is given in Fig.~\ref{fig:synthetic_demo}). Top: The embedding in latent space at different epoch. Bottom: Corresponding similarity distributions of the embeddings. The distribution of inter- and intra-class similarities of two nearest clusters become localized (bottom-right) after training. \vspace{-15pt}}
    \label{fig:embedding_visulaization}
\end{figure*}

\vspace{-2pt}
\subsection{Clustering in the Latent Space}

Let $\mathbf{C}$ be the centers of the clusters which are learnable parameters learned by our clustering model
and $\mathbf{H}_{i}$ 
be the $i$-th low-dimensional sample. 
Attracting intra-cluster samples to the same centroid is equivalent to increasing their cluster assignment confidences. 
Inspired by 
\cite{maaten2008visualizing} that transform distances to joint probability among samples, 
we compute a cluster assignment probability 
$\mathbf{Q}$ as
\begin{equation}
\small
    \label{eq:q_distribution}
    \mathbf{Q}_{ij} = \frac{(1 + || \mathbf{H}_{i} - \mathbf{C}_{j}||^{2} )^{-1}}{\sum_{k} (1 + || \mathbf{H}_{i} - \mathbf{C}_{k}||^{2})^{-1}}
\end{equation}
where $\mathbf{C}_{j}$ is the $j$-th cluster center. 
Each row in $\mathbf{Q}$ indicates the probability of that sample belongs to a particular cluster. 
During training, to iteratively increase the cluster assignment 
confidence, 
we 
first calculate a target cluster assignment matrix $\mathbf{P}$ from $\mathbf{Q}$, and then minimize the clustering loss $\mathcal{L}_{clu}$, i.e., the Kullback–Leibler (KL) divergence between the cluster assignment matrix $\mathbf{Q}$ and $\mathbf{P}$:  
\begin{equation}
\small
\label{eq:intra_reg}
    \mathcal{L}_{clu} = KL(\mathbf{P}||\mathbf{Q}), ~\text{where }\mathbf{P}_{ij} = \frac{\mathbf{Q}_{ij}^{2}/\sum_{i^\prime}\mathbf{Q}_{i^{\prime}j}}{\sum_{k}(\mathbf{Q}_{ik}^{2}/\sum_{i^\prime}\mathbf{Q}_{i^{\prime}k})}. 
\end{equation}

Intuitively, minimizing $\mathcal{L}_{clu}$ will make the largest cluster assignment probability of each row $\mathbf{Q}_i$ more dominant as it becomes more similar to $\mathbf{P}_i$.

Finally, the objective function to achieve VCC 
combines $\mathcal{L}_{bps}$, $\mathcal{L}_{c}$, $\mathcal{L}_{e}$ and $\mathcal{L}_{clu}$ as
{ \begin{eqnarray}
\small
    \label{eq:ecc_loss}
    \mathcal{L} =  \mathcal{L}_{bps} + \mathcal{L}_{c} + \mathcal{L}_{e} + \beta\mathcal{L}_{clu}.
\end{eqnarray}}%

Since the clustering process should begin after the boundary points are (at least naively) separated, 
here, we introduce a dynamic parameter $\beta$ as an increasing function of epoch index $E$ (i.e., $\beta = \gamma E$) to gradually enroll $\mathcal{L}_{clu}$. 
Minimizing $\mathcal{L}$ will first focus more on data embedding and then move on to clustering with $\beta$ to 
achieve highly compact and sparse clusters in the low-dimensional latent space. 
This $\mathcal{L}$ can be minimized using gradient-based backpropagation.

Of course, these individual losses can be weighted, but it requires introduction of complicated hyperparameters. 
We have empirically confirmed that the framework works well with even contributions from the losses, and their behaviors are 
discussed in the ablation study in the Experiment section.

\vspace{-5pt}
\section{Experiments}
\label{sec:experiments}
In this section, we 
evaluate the performance of our clustering framework, i.e., VCC, on several high-dimensional large public datasets by comparing it with various state-of-the-art (SOTA) clustering algorithms.

\vspace{-2pt}
\subsection{Experimental Setup}
\vspace{-2pt}
\paragraph{Datasets.} To evaluate performances of different clustering methods, five popular (image and non-image) public datasets were used: 
REUTERS-10K~\cite{idec}, MNIST, MNIST-test, USPS, and Fashion-MNIST. 
Each dataset contains several classes 
with ground truths. 
The datasets are summarized in TABLE~\ref{tab:datasets}.

\begin{table}[tbh]
\caption{\footnotesize Summary of Datasets. \vspace{-5pt}}
\centering
\scalebox{.7}{
\begin{tabular}{ lccc }
\toprule
 Dataset & \# Samples & \# Classes & Sample Dimension \\
 \midrule
 REUTERS-10K & 10,000 & 4 & 2000 \\
 MNIST & 70,000 & 10 &  $28 \times 28$ \\ 
  MNIST-test & 10,000 & 10 &  $28 \times 28$ \\ 
  USPS & 9,298 & 10 &  $16 \times 16$ \\ 
  Fashion-MNIST & 10,000 & 10 &  $28 \times 28$ \\
  \bottomrule
\end{tabular}
\label{tab:datasets}}
\vspace{-17pt}
\end{table}

\vspace{-2pt}
\paragraph{Baseline Methods.} 
We adopt a broad range of shallow to deep clustering methods for comparison shown in TABLE~\ref{tab:clustering_performance}. All the shallow methods are listed in the top panel. Based on the usage of the convolutional technique,  deep methods are divided into two categories: 
without convolution (i.e., middle panel) and 
with convolution (i.e., bottom panel).

\vspace{-2pt}
\paragraph{Evaluation Metrics.} 
To evaluate the performance of various clustering algorithms, we adopt the two most common evaluation metrics (i.e., Normalized Mutual Information (NMI) and Clustering Accuracy (ACC)) in our experiments. Different from ACC which is computed by finding the best match between cluster assignment and target labels, NMI has the capability of capturing the similarity between cluster assignment and target labels \cite{xu2003document}.

\begin{table*}[!t]
\vspace{-10pt}
\centering
\caption{
\footnotesize Clustering performances on public datasets. Top panel: Shallow methods, Middle panel: Deep methods, Bottom panel: Methods with Convolution (or image features).  The results of baselines were taken from \cite{ren2020deep,yang2019deep}. ``--" and ``${*}$"mean the results are not available or obtained by running provided codes, respectively. 
Pre-trained VGG \cite{Simonyan15} is used to extract features from images if convolution is separately needed. \vspace{-5pt}
}
\vspace{-2pt}
\label{tab:clustering_performance}
\scalebox{.78}{
\begin{tabular}{ l|c|lllllllllllllllll }
\toprule
& w/& \multicolumn{2}{c}{MNIST} & \multicolumn{2}{c}{MNIST-test}  & \multicolumn{2}{c}{USPS} &  \multicolumn{2}{c}{Fashion-MNIST} & \multicolumn{2}{c}{REUTERS-10K}  \\ \cmidrule{3-12}
& Conv. & ACC & NMI & ACC & NMI & ACC & NMI & ACC & NMI & ACC & NMI \\ \midrule
$k$-means & \xmark & 0.500 & 0.534 & 0.501 & 0.547 & 0.450 & 0.460 & 0.476 & 0.512 & 0.516 & 0.309$^{*}$ \\ 
DBSCAN & \xmark &  --  &  --  &  0.114  &  0  &  0.167  &  0  & 0.100 & 0 & 0.403 & 0.003 \\ 
DenPeak & \xmark & --  &  --  &  0.357  &  0.399  &  0.390  &  0.433 & 0.344 & 0.398 & -- & --  \\ 
N-Cut~\cite{ncut} & \xmark & 0.411  &  0.327  &  0.753  &  0.304 &  0.675  &  0.314  & -- & -- &  -- & -- \\ 
LDMGI~\cite{ldmgi} & \xmark & 0.802  &  0.842  &  0.811  &  0.847  &  0.563  &  0.580  & -- & -- & -- & -- \\ 
SC-ST~\cite{stsc} & \xmark & 0.416  &  0.311  &  0.756  &   0.454  &  0.726  &  0.308  & -- & -- & -- & -- \\ 
SC-LS~\cite{scls} & \xmark & 0.706  &  0.714  &  0.756  &  0.740  &  0.681  &  0.659  & -- & -- & -- & -- \\  \midrule
DEC~\cite{xie2016unsupervised} & \xmark & 0.849  &  0.816  &  0.856  &  0.830  &  0.758  &  0.769 & 0.591 & 0.618  & 0.737 & 0.497 \\ 
IDEC~\cite{idec} & \xmark & 0.881  &  0.867  &  0.846  &  0.802  &  0.759  &  0.777  & 0.523 & 0.600 & 0.756 & 0.498 \\ 
DKM~\cite{fard2020deep} & \xmark & 0.840  &  0.796  &  --  &  --  &  0.757  &  0.776  & -- & --  & -- & -- \\
DCN~\cite{dcn} & \xmark & 0.830  &  0.810  &  0.802  &  0.786  &  0.688  &  0.683  & -- & -- & -- & -- \\ 
$\text{UMAP}^{kmeans}$ & \xmark &  0.759 &  0.713  &  0.848  &  0.789  &  0.764  & 0.789   & 0.580 & 0.569 & \textbf{0.781} & \textbf{0.571}  \\ 
{\textbf{VCC} (ours)} & \xmark & 0.964  &  0.921  &  0.946  &  0.889  &  0.970  &  0.937  & \textbf{0.670}  & 0.656  & \textbf{0.812 } & \textbf{0.605 }  \\\midrule\midrule
$k\text{-means}_{VGG}$ & \cmark & 0.804  &  0.682  &  0.822  &  0.721  &  0.792  &  0.837  &  0.552 &  0.598 & -- & --  \\ 
JULE~\cite{jule} & \cmark & 0.964  &  0.913  &  0.961  &  0.915  &  0.950  &  0.913  & -- & -- & -- & --  \\ 
DEPICT & \cmark & 0.965  &  0.917  &  0.963  &  0.915  &  0.964  &  0.927  & -- & -- & -- & --  \\ 
ConvDEC & \cmark & 0.940  &  0.916  &  0.861  &  0.847 &  0.784 &  0.820  & 0.514 & 0.588 & -- & -- \\ 
ConvDEC-DA & \cmark & \textbf{0.985}  &  \textbf{0.961}  &  0.955  &  \textbf{0.949}  &  0.970  &  \textbf{0.953}  & 0.570 & 0.632 & -- & -- \\ 
DDC~\cite{ren2020deep} & \cmark & 0.965  &  0.932  &  0.965  &  0.916  &  0.967  &  0.918  & 0.619 & \textbf{0.682}  & -- & -- \\ 
DDC-DA & \cmark  & 0.969  &  0.941  &  \textbf{0.970}  &  0.927  &  0.977  &  0.939  & 0.609 & 0.661  & -- & -- \\
$\text{UMAP}^{kmeans}_{VGG}$ & \cmark & 0.909  & 0.867   &  0.962  &  0.916  &  \textbf{0.978}  & 0.945   & 0.593  & 0.605 & -- & -- \\ 
{\textbf{$\text{VCC}_{VGG}$} (ours)} & \cmark & \textbf{0.983 }  & \textbf{0.971 }   &  \textbf{0.982}  &  \textbf{0.969}  &  \textbf{0.981 }  &  \textbf{0.969 }  & \textbf{0.665}  & \textbf{0.703 } & -- & -- \\
\bottomrule
\end{tabular}
}
\vspace{-17pt}
\end{table*}

\vspace{-2pt}
\paragraph {Implementation Details.} The input data is projected into a low-dimensional latent space utilizing an MLP (with layers: 500, 500, 2000) which is widely used in other deep clustering methods. Technically, $\mathcal{H}$ can be in any dimension, but 2-D was sufficient to get good clustering results and perceptually interpretable visualization. When constructing Latent Graph $\mathcal{G}_{lg}$, the number of nearest neighbors is set to 10, and the distance metric is set to Euclidean by default.
Specifically, the number of nearest number for REUTERS-10K is set to 70. For the MLP, we use Stochastic Gradient Descent (SGD) as the optimizer with learning rate $0.01$, momentum $0.9$, and weight decay $0.0005$ and the batch size is set to 200. The $\gamma$ used for $\beta$ was $0.01$.
All the experiments are implemented in an Ubuntu environment equipped with RTX8000 GPUs.

\vspace{-5pt}
\subsection{Results and Discussions}

\vspace{-2pt}
\paragraph{Clustering on Public Datasets.}  
TABLE~\ref{tab:clustering_performance} 
compares the clustering performances of baseline methods and VCC on various datasets. 
Top-2 algorithms on each dataset are highlighted in bold. 
TABLE~\ref{tab:clustering_performance} shows 
that most deep methods perform better than shallow models. 
Among the deep methods,
comparing DEC and ConvDEC clearly tells that the convolution results in an increase in clustering performance for image data. 
Moreover, comparisons of DDC vs. DDC-DA and ConvDEC vs. ConvDEC-DA show 
that Data Augmentation (DA) provides an improvement as well. 

{\em Without Convvolution:} Shallow methods mostly did not perform well on these high-dimensional data. 
Deep methods that perform dimension reduction and $k$-means with UMAP yielded reasonable results. 
On the other hand, VCC achieved the best performance in both ACC and NMI on all five datasets 
across image and non-image data. 
VCC's performances on image data were also comparable with or sometimes even better than the results from the baselines 
that utilize convolution to extract effective image representations. 
These results tell that VCC is able to learn very effective representation of the data regardless of its modality. 

{\em With image features:} 
The VCC with VGG features 
achieved at least top-2 ACC ($\sim$$0.98$) and NMI ($\sim$$0.95$) on MNIST and USPS datasets {\em even without} DA, 
and only ConvDEC with DA yielded slightly better result than VCC on the accuracy. 
Even for 
more complicated image data 
(i.e., Fashion-MNIST), 
VCC achieved pretty good ACC (0.665) and NMI (0.703) which were the best among all convolutional models. 
One notable result is that VCC without VGG features yielded the highest accuracy on the Fashion-MNIST data. 
This may be because, while VGG features are effective, their dimensions are too high compared to the original image size. 
As VCC finds an effective embedding even for images, this result is not very surprising. 

\vspace{-3pt}
\paragraph{Model Analysis.}

Fig.~\ref{fig:embedding_visulaization} shows the training process of VCC using MNIST-test data in 2-D space which is the low-dimensional space that the model learns. 
Given the original data represented in 2-D as in Fig.~\ref{fig:synthetic_demo} (a), it can be seen that the data samples are gradually being separated as the training progresses as in Fig.~\ref{fig:embedding_visulaization} (a), (b) and (c). 
At the final epoch, individual clusters are compactly clustered with very few falsely clustered samples. 
The bottom panel in Fig.~\ref{fig:embedding_visulaization} shows intra- and inter-class similarity distributions of the two nearest clusters (i.e., classes) in (a) to demonstrate the quality of the clusters. It is interesting to see that the intra-class similarities in both clusters (red and blue) become highly localized to small values during the optimization, and inter-class similarity distribution (purple) gets shifted to larger values. 
Such a behavior is exactly what we expected with VCC, i.e., making individual cluster compact and separating the clusters.

\begin{table}[!b]
\vspace{-15pt}
\centering
\caption{\footnotesize Comparisons with SOTA methods on challenging image datasets. VCC achieves competitive results.   \vspace{-7pt}}
\label{tab:comparison_sota}
\scalebox{.65}{
\begin{tabular}{ l|cc|cc|cc }
\toprule
 &  \multicolumn{2}{c}{CIFAR10} & \multicolumn{2}{c}{CIFAR100-20} & \multicolumn{2}{c}{STL10} \\ \cmidrule{2-7}
 & ACC & NMI   &   ACC & NMI  &   ACC & NMI  \\ \midrule
DeepCluster~\cite{caron2018deep} &  0.374 & -  & 0.189  & -  & 0.334  & -  \\
DAC~\cite{chang2017deep} &  0.522 & 0.400 & 0.238  & 0.185  & 0.470  & 0.366  \\
IIC~\cite{ji2019invariant} &  0.617 & 0.511  & 0.257  & 0.225  & 0.596  & 0.496  \\
SCAN-Loss(SimCLR)  &  0.787 & -  &   - & -  & - & -  \\
SCAN-Loss(RA)  &  0.818 & 0.712  &   0.422 & 0.441  & 0.755 & 0.654  \\ 
VCC(ours) &  0.809 & 0.698  & 0.372  & 0.426 & 0.722 & 0.616  \\
\bottomrule
\end{tabular}}
\vspace{-5pt}
\end{table}

\vspace{-5pt}
\subsection{Comparisons on Challenging Image Dataset}

TABLE~\ref{tab:comparison_sota} 
compares the performances of other recent image clustering methods 
with VCC on more complex datasets including CIFAR10 \cite{krizhevsky2009learning}, CIFAR100-20 \cite{krizhevsky2009learning} and STL10 \cite{coates2011analysis}. 
We provide these results separately as their results on MNIST, USPS, Fashion-MNIST and Reuters-10K have not been reported. 
The experimental settings follow SCAN~\cite{van2020scan} which trains and evaluates using the train and validation splits, respectively. 
The details of the parameters and the summary of these datasets are given in supplementary.

TABLE~\ref{tab:comparison_sota} shows that SCAN \cite{van2020scan} and our VCC outperform other methods with a large gap ($> 17\%$) in accuracy. 
SCAN uses is a self-supervised learning technique (SimCLR) from \cite{chen2020simple} 
and a strong data augmentation for self-labeling, i.e., RandAugment (RA) \cite{cubuk2020randaugment}. 
For fair comparisons, VCC here also adopted the same embedding from SimCLR as an input. 
While SCAN showed the best results, however when the RA was removed, 
VCC performed better than SCAN by $2.2\%$ accuracy on CIFAR10. 
It shows that VCC is a competitive framework; it can be used for non-image data and yields qualitative visualization of clusters that other methods may not provide.

\vspace{-5pt}
\subsection{Ablation Studies}

\vspace{-2pt}
\paragraph{Significance of Losses.} TABLE~\ref{tab:supp_clustering_performance} shows our ablation study on Contraction and Expansion losses. From TABLE~\ref{tab:supp_clustering_performance}, we can see that the performance of VCC drops (i.e., ACC drops from 0.94 to 0.80 and NMI decreases from 0.89 to 0.82) after removing Contraction and Expansion losses. Moreover, TABLE~\ref{tab:supp_clustering_performance} also shows that the expansion is critical in clustering accuracy 
{\em emphasizing the importance of separating boundary points}. 

\begin{table}[!b]
\vspace{-14pt}
\centering
\caption{\footnotesize VCC clustering performance comparison between settings with and without Contraction loss and Expansion loss. \vspace{-5pt}}
\label{tab:supp_clustering_performance}
\scalebox{.8}{
\begin{tabular}{ l|c|c|ll }
\toprule
& w/ & w/ & \multicolumn{2}{c}{MNIST-test}  \\ \cmidrule{2-5}
& Contraction ($\mathcal{L}_{c}$) & Expansion ($\mathcal{L}_{e}$) & ACC & NMI  \\ \midrule
\multirow{4}{*}{VCC}  & \checkmark & \checkmark &  0.946  &  0.889   \\
  & \xmark & \checkmark &  0.931  &  0.864   \\ 
  & \checkmark & \xmark &  0.810  &  0.852   \\ 
 & \xmark & \xmark & 0.801  &  0.820  \\ 
\bottomrule
\end{tabular}}
\vspace{-5pt}
\end{table}

\begin{figure}[!t]
  \begin{minipage}[!h]{.45\textwidth}
  \centering
    \begin{subfigure}{.3\textwidth}
        \centering
        \includegraphics[width=\textwidth]{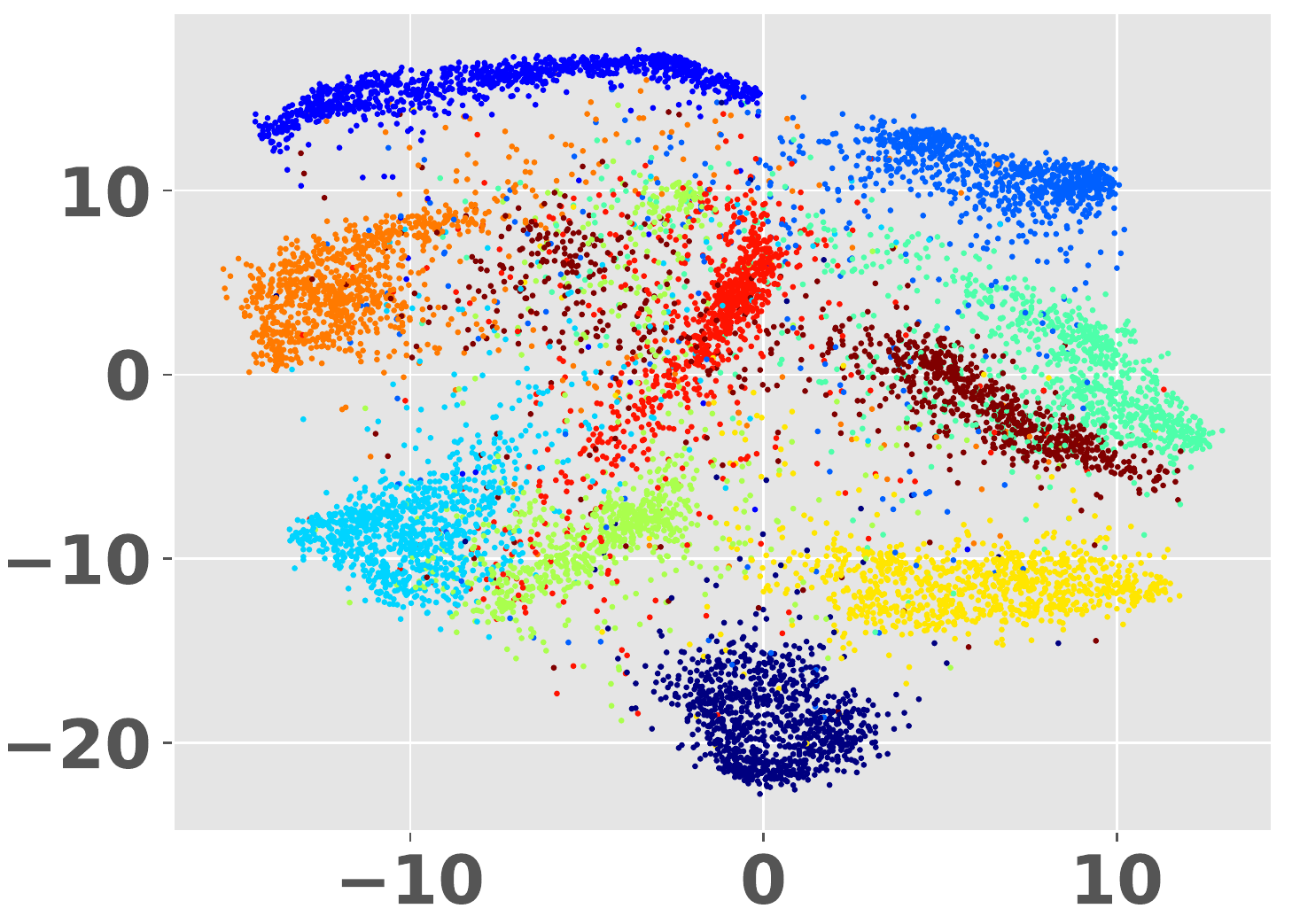}
    \end{subfigure}\hfill
    \begin{subfigure}{.3\textwidth}
        \centering
        \includegraphics[width=\textwidth]{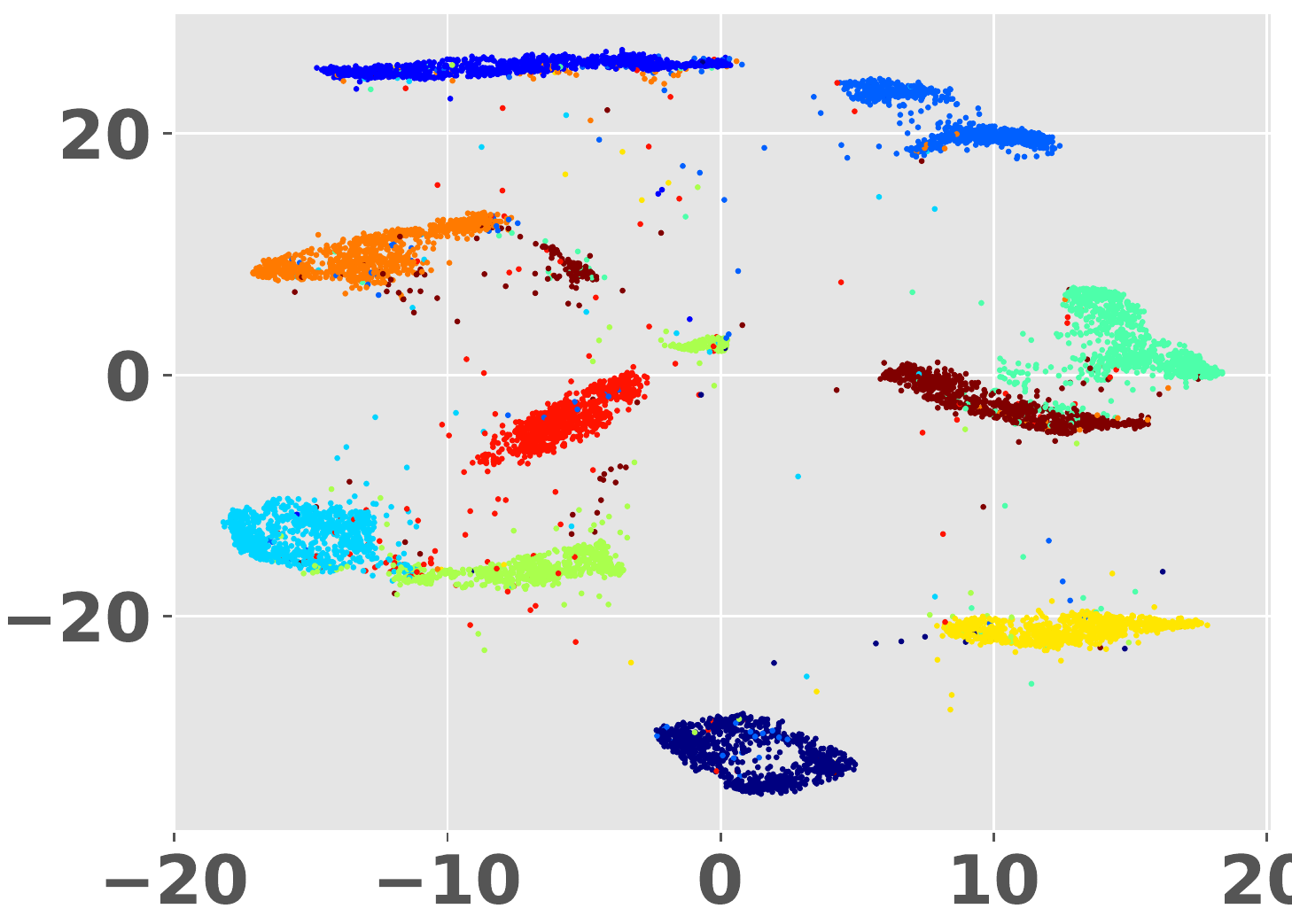}
    \end{subfigure}\hfill
    \begin{subfigure}{.3\textwidth}
        \centering
        \includegraphics[width=\textwidth]{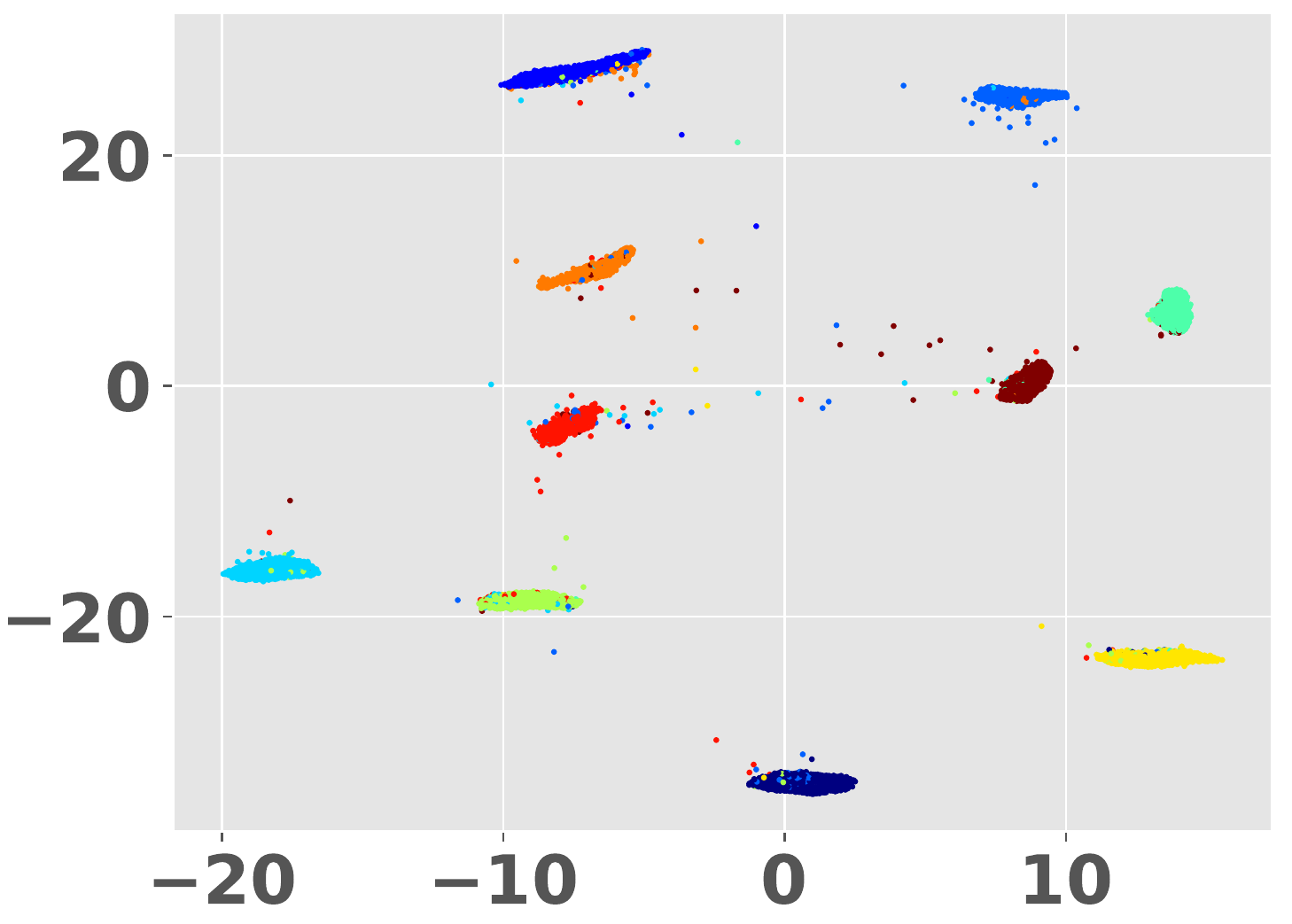}
    \end{subfigure}
    \begin{subfigure}{.3\textwidth}
        \centering
        \includegraphics[width=\textwidth]{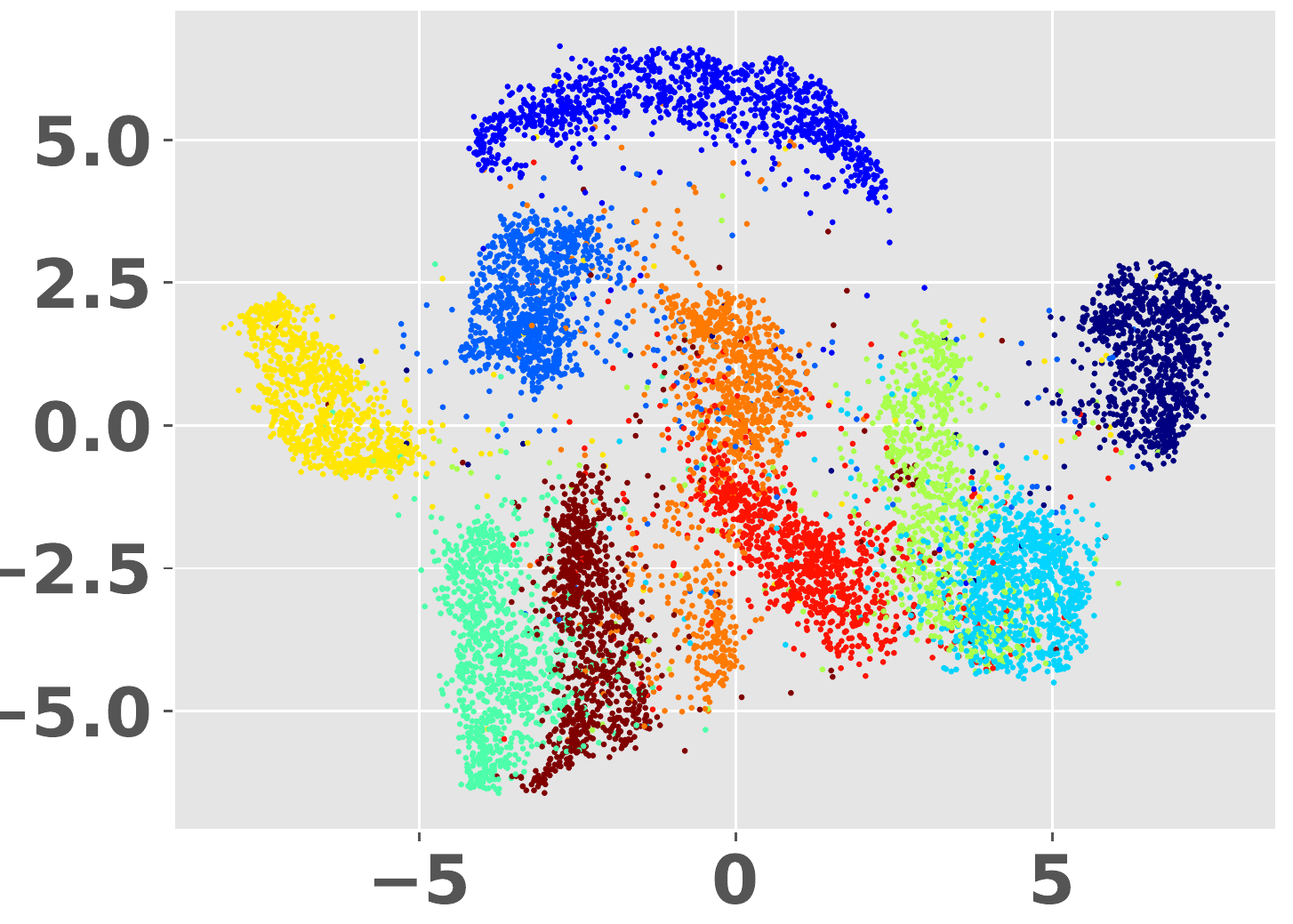}
    \end{subfigure}\hfill
    \begin{subfigure}{.3\textwidth}
        \centering
        \includegraphics[width=\textwidth]{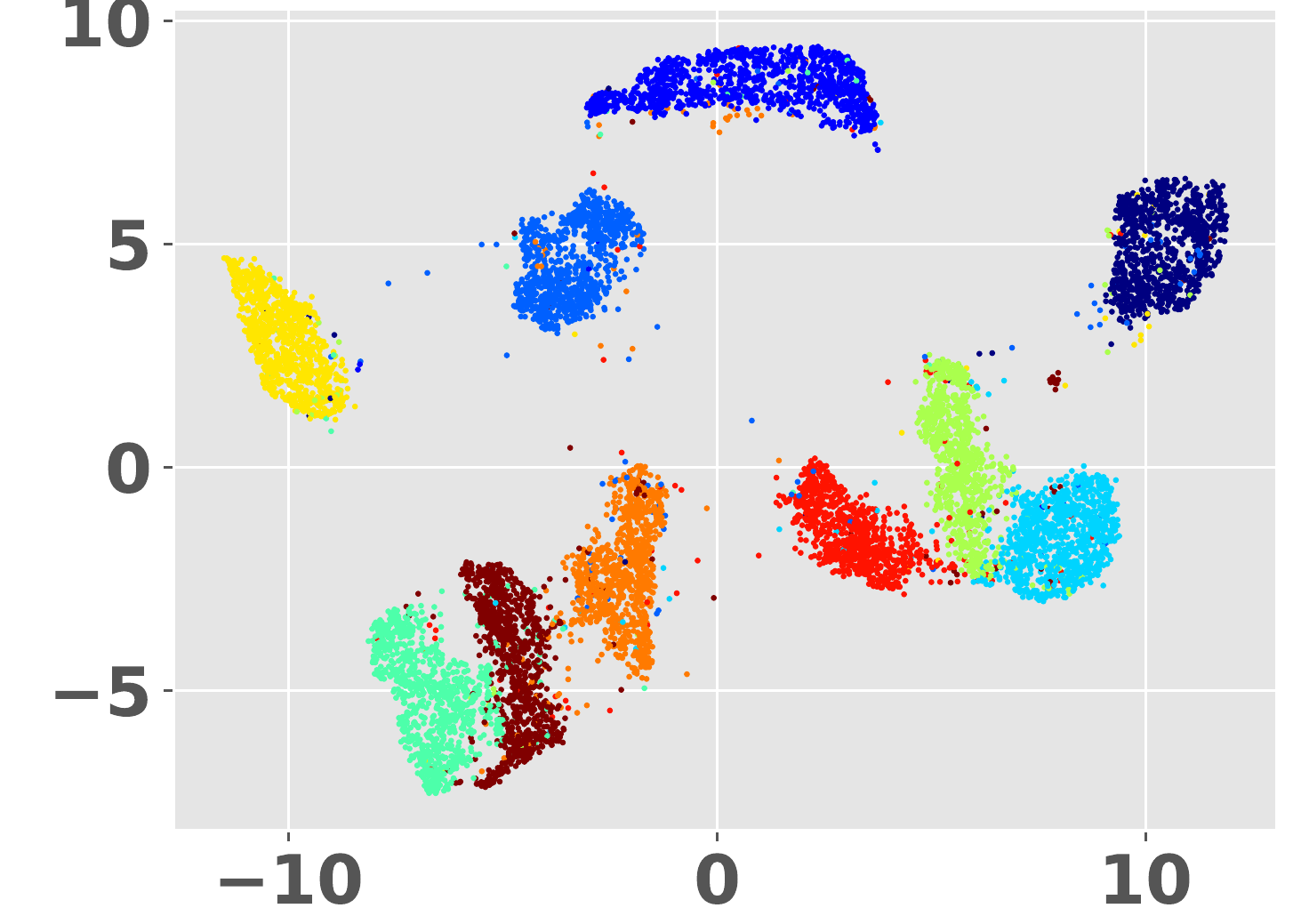}
    \end{subfigure}\hfill
    \begin{subfigure}{.3\textwidth}
        \centering
        \includegraphics[width=\textwidth]{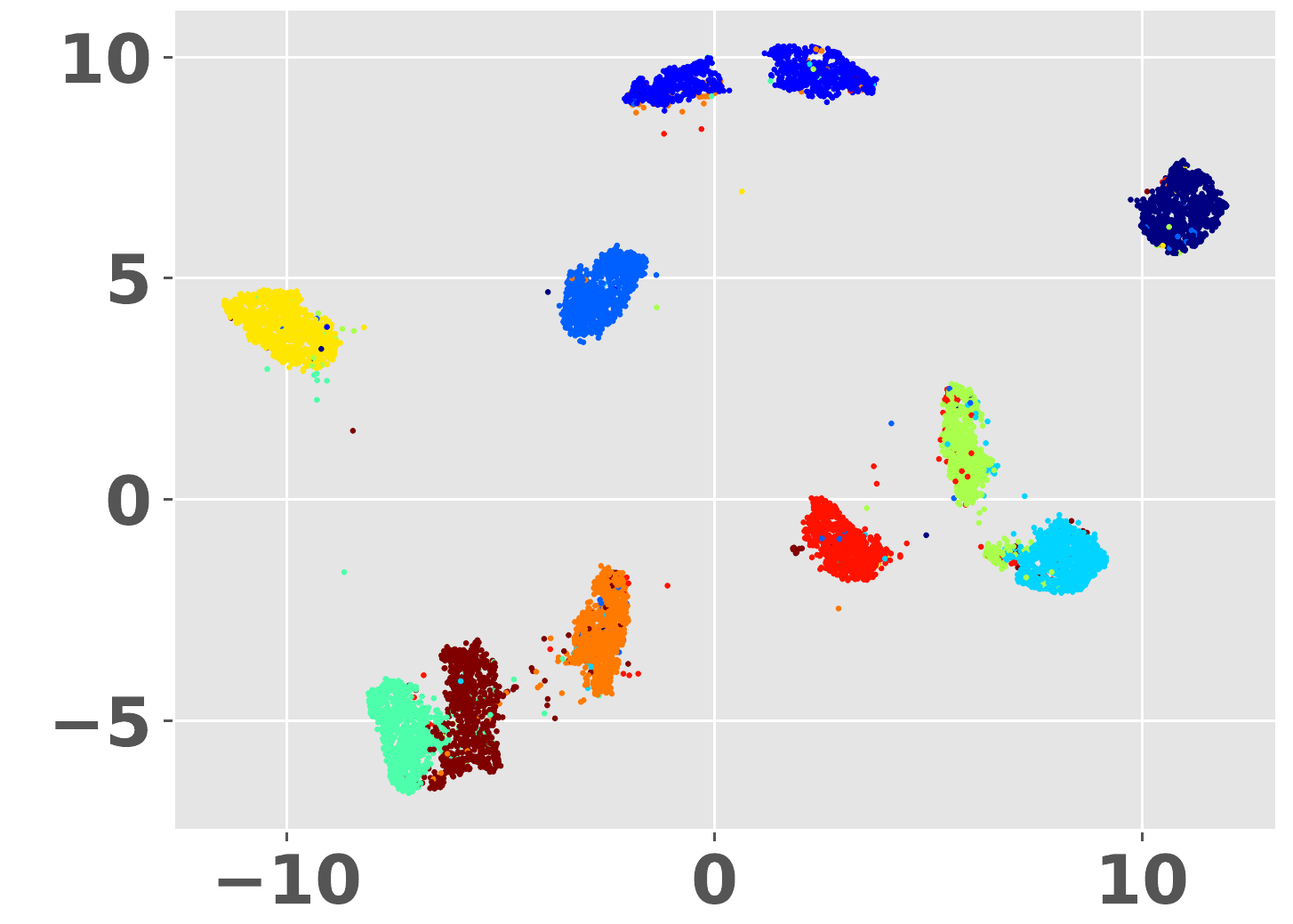}
    \end{subfigure}
    \begin{subfigure}{.3\textwidth}
        \centering
        \includegraphics[width=\textwidth]{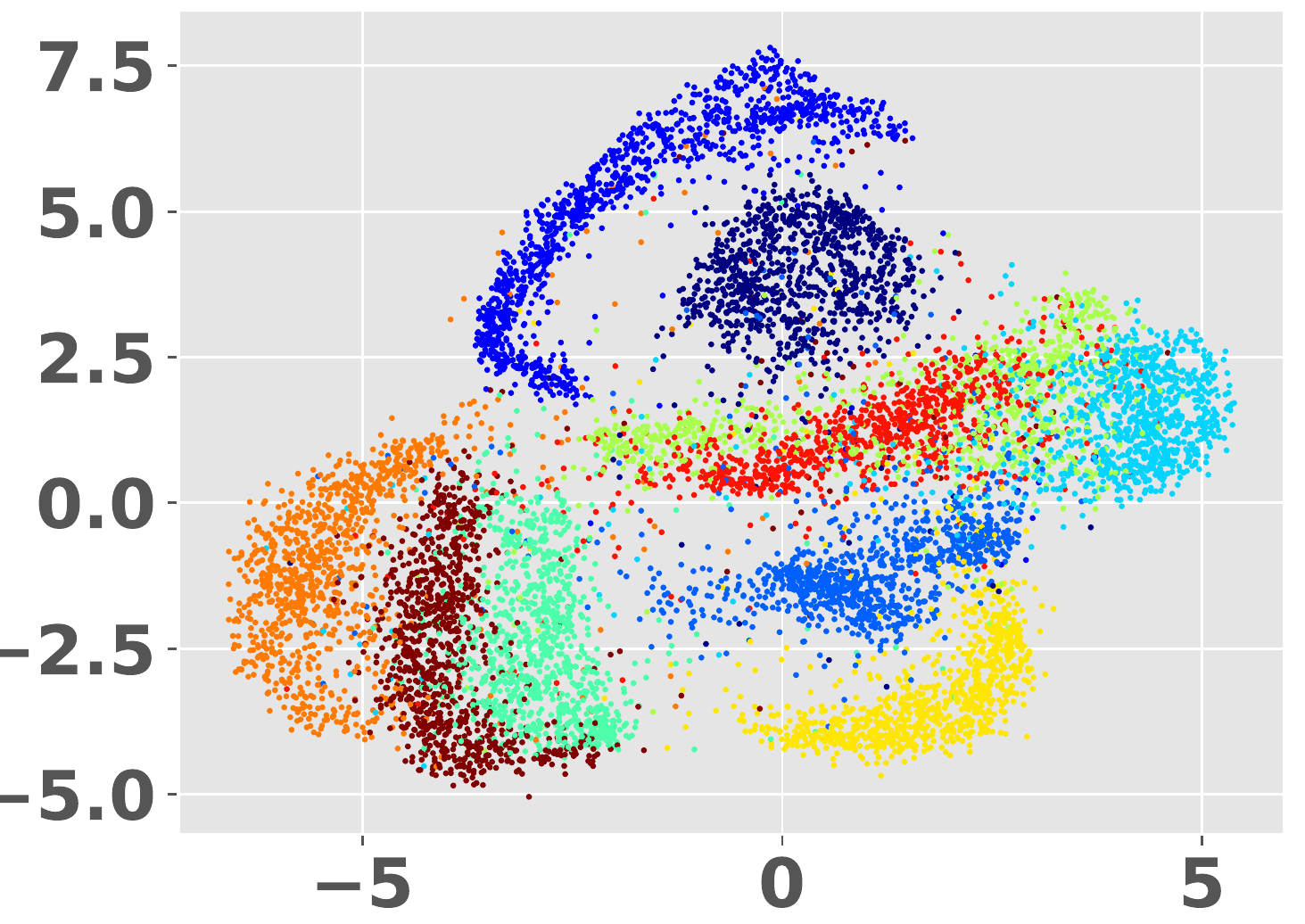}
        \caption{\footnotesize Beginning.}
    \end{subfigure}\hfill
    \begin{subfigure}{.3\textwidth}
        \centering
        \includegraphics[width=\textwidth]{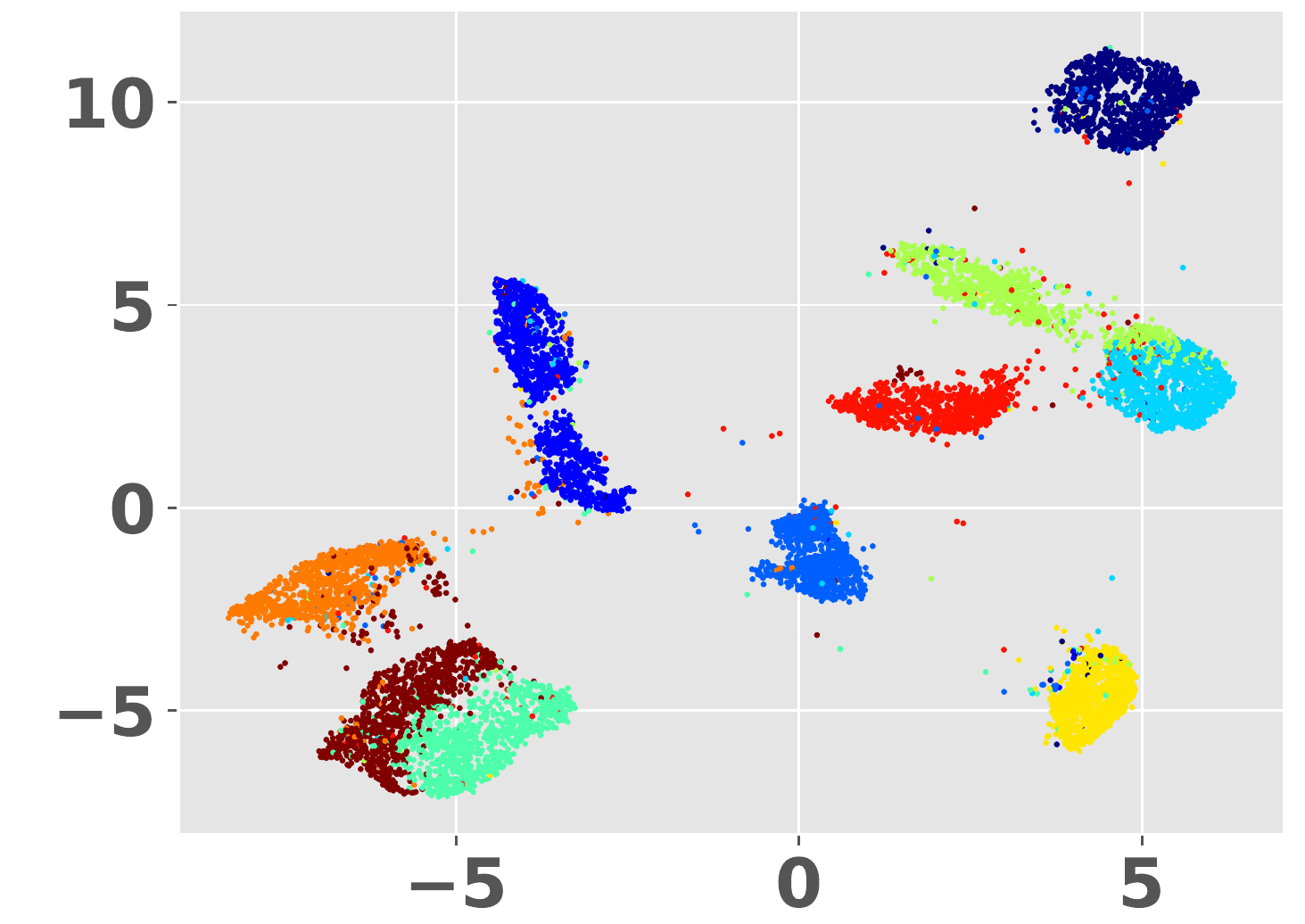}
        \caption{\footnotesize Middle.}
    \end{subfigure}\hfill
    \begin{subfigure}{.3\textwidth}
        \centering
        \includegraphics[width=\textwidth]{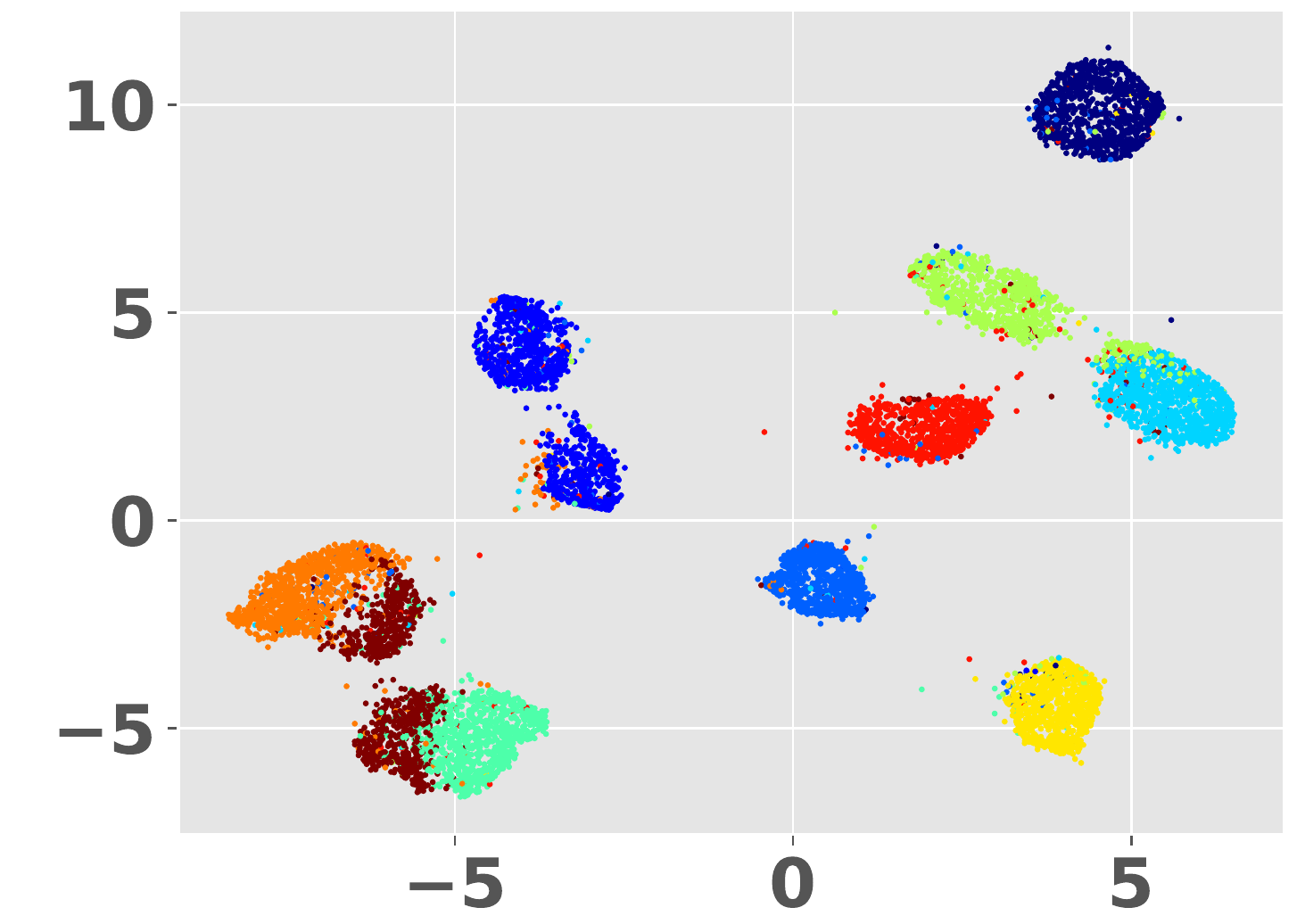}
        \caption{\footnotesize Final.}
    \end{subfigure}
    \end{minipage}
    \vspace{-8pt}
    \caption{\footnotesize Visualization of VCC's clustering assignment performance on MNIST-test data throughout training (the input embedding is given in Fig.~1 in the main manuscript) without using Contraction loss $\mathcal{L}_{c}$ and Expansion loss $\mathcal{L}_{e}$. Top: VCC without $\mathcal{L}_{c}$. Middle: VCC without $\mathcal{L}_{e}$. Bottom: VCC without $\mathcal{L}_{c}$ and $\mathcal{L}_{e}$. 
    \vspace{-10pt}
    }
    \label{fig:supp_vcc_embedding_wo_ECEE}
\end{figure}

Fig.~\ref{fig:supp_vcc_embedding_wo_ECEE} illustrates the embedding of VCC in the latent space at different stages after removing the Contraction loss or Expansion loss. The good embedding result at the beginning epoch indicates that the boundary points separation loss converges fast, which benefits from the sampling strategy based on edges. However, when comparing it with the middle epoch and final epoch embeddings, we can see that separating boundary points alone does not
guarantee a good clustering performance. The failure cases in the middle/bottom row of Fig.~\ref{fig:supp_vcc_embedding_wo_ECEE}, i.e., merged clusters, show that contraction and expansion need to be properly performed besides separating boundary points. 
More interestingly, the comparison between the embedding results at the top and middle rows demonstrates the behavior of VCC shown in  
TABLE~\ref{tab:supp_clustering_performance}, i.e., 
the expansion process can separate different clusters far away to reduce the inter-similarities among clusters dramatically.

\begin{figure}[!tb]
\vspace{-5pt}
  \centering
    \includegraphics[width=.75\linewidth]{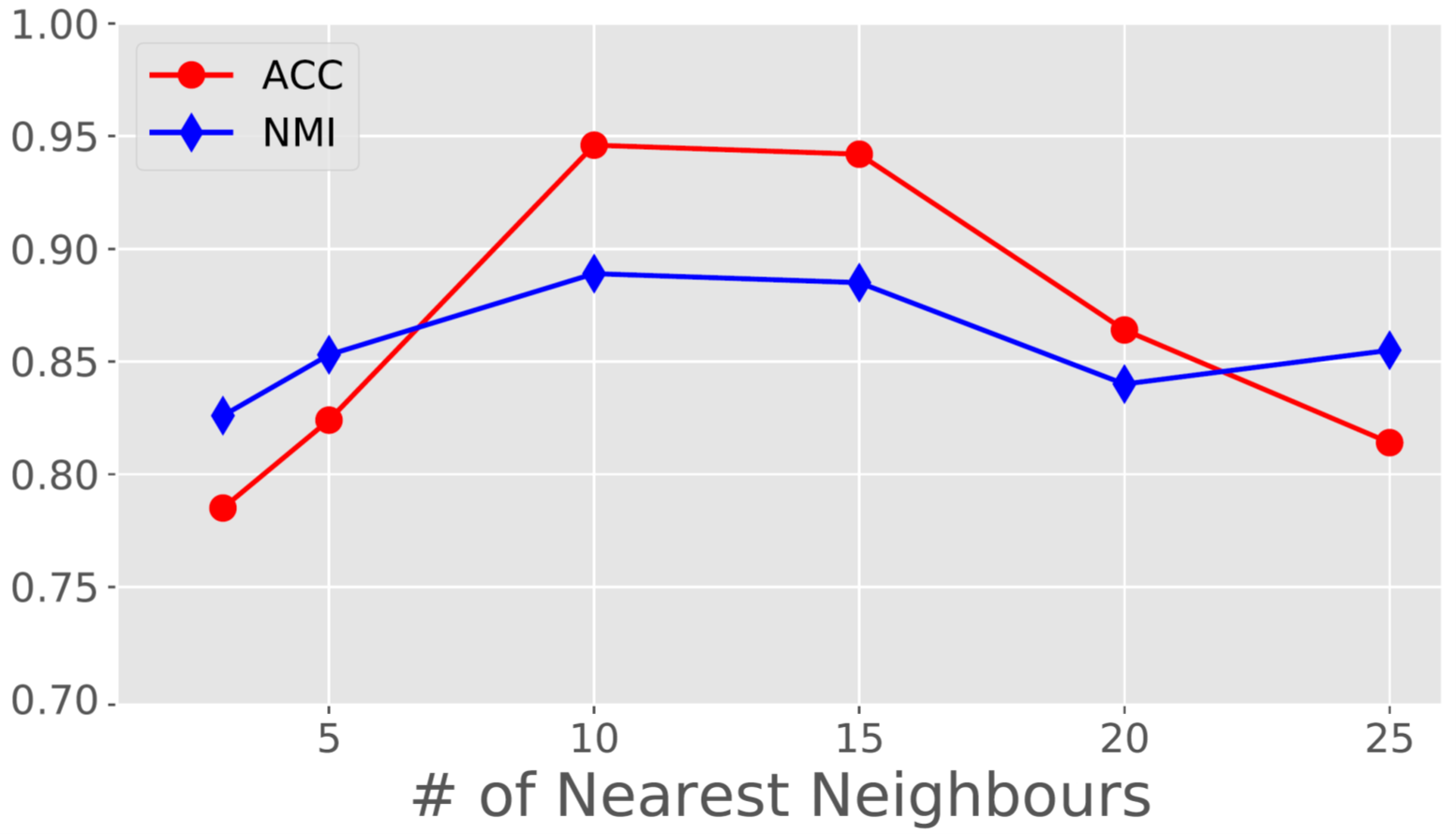}
    \vspace{-8pt}
    \caption{\footnotesize Performance changes with different numbers of nearest neighbors $M$. \vspace{-5pt}}
    \vspace{-15pt}
    \label{fig:adlation_study_acc_nmi}
\end{figure}

\vspace{-2pt}
\paragraph{Nearest Neighbors.}  As discussed in the method Section, $k$NN affects VCC via boundary points. 
We therefore perform an ablation study on the number of neighbors used for VCC using MNIST-test dataset without any image features such as VGG. 
The result is well summarized in Fig.~\ref{fig:adlation_study_acc_nmi}. 
The number of neighbors $M$ are changed as 3, 5, 10, 15, 20, 25 to obtain 
accuracies and NMI in Fig.~\ref{fig:adlation_study_acc_nmi}. 
At the beginning, both the accuracy and NMI increase as $M$ increases, however, they start to drop after $M=15$ especially in the accuracy. 
This can be explained similarly to the {\em bias-variance trade-off} of $k$-nearest neighbor classifiers \cite{domingos2000unified}. 
As $M$ increases, the bias in $k$NN graph increases, and it results in the drop in the precision of clustering of VCC. 
Such a behavior was expected, and the same can happen in any algorithms where variants of $k$NN are adopted. 

\vspace{-2pt}
\paragraph{Limitations.} 
VCC may not be the best choice for datasets with predefined relationships (e.g., graphs), where the relationships between data points are determined by both the underlying manifold structure and its own explicit structure. Moreover, the capability of $\mathcal{G}_{lg}$ to capture local and global structures hidden in the dataset also affects the clustering performance. Fortunately, this limitation can be eased by the rapid improvement of representation learning (e.g., self-supervised learning).

\vspace{-7pt}
\section{Conclusion}
In this paper, we proposed a 
novel end-to-end clustering algorithm, i.e., VCC, for general datasets by making full use of structure information within the data. 
The key of VCC is to first separate samples near the boundary of clusters 
given observations on variations of similarities among neighboring samples. 
The algorithm finds a low-dimensional latent space where data samples within the same group form a highly compact cluster 
and the different clusters become distinct from each other. 
Experiments comparing VCC with other state-of-the-art baseline methods on public datasets demonstrate that VCC outperforms in general, 
and it is able to learn highly effective embeddings of data even for image data without convolution operation. 
The code will be publicly available via open-source project platforms.

{\small 
\bibliography{aaai22.bib}

\begin{thebibliography}{39}
\providecommand{\natexlab}[1]{#1}

\bibitem[{Bolin et~al.(2014)Bolin, Edwards, Finch, and
  Cassady}]{bolin2014applications}
Bolin, J.~H.; Edwards, J.~M.; Finch, W.~H.; and Cassady, J.~C. 2014.
\newblock Applications of cluster analysis to the creation of perfectionism
  profiles: a comparison of two clustering approaches.
\newblock \emph{Frontiers in psychology}, 5: 343.

\bibitem[{Caron et~al.(2018)Caron, Bojanowski, Joulin, and
  Douze}]{caron2018deep}
Caron, M.; Bojanowski, P.; Joulin, A.; and Douze, M. 2018.
\newblock Deep clustering for unsupervised learning of visual features.
\newblock In \emph{ECCV}, 132--149.

\bibitem[{Chang et~al.(2017)Chang, Wang, Meng, Xiang, and Pan}]{chang2017deep}
Chang, J.; Wang, L.; Meng, G.; Xiang, S.; and Pan, C. 2017.
\newblock Deep adaptive image clustering.
\newblock In \emph{ICCV}, 5879--5887.

\bibitem[{Chen et~al.(2020)Chen, Kornblith, Norouzi, and
  Hinton}]{chen2020simple}
Chen, T.; Kornblith, S.; Norouzi, M.; and Hinton, G. 2020.
\newblock A simple framework for contrastive learning of visual
  representations.
\newblock In \emph{ICML}, 1597--1607. PMLR.

\bibitem[{Chen and Cai(2011)}]{scls}
Chen, X.; and Cai, D. 2011.
\newblock Large scale spectral clustering with landmark-based representation.
\newblock In \emph{AAAI}. Citeseer.

\bibitem[{Coates, Ng, and Lee(2011)}]{coates2011analysis}
Coates, A.; Ng, A.; and Lee, H. 2011.
\newblock An analysis of single-layer networks in unsupervised feature
  learning.
\newblock In \emph{AISTATS}, 215--223.

\bibitem[{Costa and Ortale(2020)}]{costa2020document}
Costa, G.; and Ortale, R. 2020.
\newblock Document clustering meets topic modeling with word embeddings.
\newblock In \emph{ICDM}, 244--252. SIAM.

\bibitem[{Cubuk et~al.(2020)Cubuk, Zoph, Shlens, and Le}]{cubuk2020randaugment}
Cubuk, E.~D.; Zoph, B.; Shlens, J.; and Le, Q.~V. 2020.
\newblock Randaugment: Practical automated data augmentation with a reduced
  search space.
\newblock In \emph{CVPR Workshop}, 702--703.

\bibitem[{Ding and He(2004)}]{ding2004k}
Ding, C.; and He, X. 2004.
\newblock K-nearest-neighbor consistency in data clustering: incorporating
  local information into global optimization.
\newblock In \emph{ACM symposium on Applied computing}, 584--589.

\bibitem[{Domingos(2000)}]{domingos2000unified}
Domingos, P. 2000.
\newblock A unified bias-variance decomposition.
\newblock In \emph{ICML}, 231--238.

\bibitem[{Ester et~al.(1996)Ester, Kriegel, Sander, Xu et~al.}]{dbscan}
Ester, M.; Kriegel, H.-P.; Sander, J.; Xu, X.; et~al. 1996.
\newblock A density-based algorithm for discovering clusters in large spatial
  databases with noise.
\newblock In \emph{KDD}, volume 96 (34), 226--231.

\bibitem[{Fard, Thonet, and Gaussier(2020{\natexlab{a}})}]{fard2020deep}
Fard, M.~M.; Thonet, T.; and Gaussier, E. 2020{\natexlab{a}}.
\newblock Deep k-means: Jointly clustering with k-means and learning
  representations.
\newblock \emph{Pattern Recognition Letters}.

\bibitem[{Fard, Thonet, and Gaussier(2020{\natexlab{b}})}]{fard2020seed}
Fard, M.~M.; Thonet, T.; and Gaussier, E. 2020{\natexlab{b}}.
\newblock Seed-Guided Deep Document Clustering.
\newblock In \emph{European Conference on Information Retrieval}, 3--16.
  Springer.

\bibitem[{Ghasedi~Dizaji et~al.(2017)Ghasedi~Dizaji, Herandi, Deng, Cai, and
  Huang}]{ghasedi2017deep}
Ghasedi~Dizaji, K.; Herandi, A.; Deng, C.; Cai, W.; and Huang, H. 2017.
\newblock Deep clustering via joint convolutional autoencoder embedding and
  relative entropy minimization.
\newblock In \emph{ICCV}, 5736--5745.

\bibitem[{Guo et~al.(2017)Guo, Gao, Liu, and Yin}]{idec}
Guo, X.; Gao, L.; Liu, X.; and Yin, J. 2017.
\newblock Improved deep embedded clustering with local structure preservation.
\newblock In \emph{IJCAI}, 1753--1759.

\bibitem[{Guo et~al.(2019)Guo, Liu, Zhu, Zhu, Li, Xu, and
  Yin}]{guo2019adaptive}
Guo, X.; Liu, X.; Zhu, E.; Zhu, X.; Li, M.; Xu, X.; and Yin, J. 2019.
\newblock Adaptive self-paced deep clustering with data augmentation.
\newblock \emph{IEEE Transactions on Knowledge and Data Engineering}, 32(9):
  1680--1693.

\bibitem[{Huang, Gong, and Zhu(2020)}]{huang2020deep}
Huang, J.; Gong, S.; and Zhu, X. 2020.
\newblock Deep semantic clustering by partition confidence maximisation.
\newblock In \emph{CVPR}, 8849--8858.

\bibitem[{Ji, Henriques, and Vedaldi(2019)}]{ji2019invariant}
Ji, X.; Henriques, J.~F.; and Vedaldi, A. 2019.
\newblock Invariant information clustering for unsupervised image
  classification and segmentation.
\newblock In \emph{ICCV}, 9865--9874.

\bibitem[{Kanezaki(2018)}]{kanezaki2018unsupervised}
Kanezaki, A. 2018.
\newblock Unsupervised image segmentation by backpropagation.
\newblock In \emph{ICASSP}, 1543--1547. IEEE.

\bibitem[{Krizhevsky, Hinton et~al.(2009)}]{krizhevsky2009learning}
Krizhevsky, A.; Hinton, G.; et~al. 2009.
\newblock Learning multiple layers of features from tiny images.
\newblock \emph{technical report}.

\bibitem[{Maaten and Hinton(2008)}]{maaten2008visualizing}
Maaten, L. v.~d.; and Hinton, G. 2008.
\newblock Visualizing data using t-SNE.
\newblock \emph{JMLR}, 9(Nov): 2579--2605.

\bibitem[{MacQueen et~al.(1967)}]{kmeans}
MacQueen, J.; et~al. 1967.
\newblock Some methods for classification and analysis of multivariate
  observations.
\newblock In \emph{Berkeley symposium on mathematical statistics and
  probability}, volume 1(14), 281--297. Oakland, CA, USA.

\bibitem[{McInnes, Healy, and Melville(2018)}]{mcinnes2018umap}
McInnes, L.; Healy, J.; and Melville, J. 2018.
\newblock Umap: Uniform manifold approximation and projection for dimension
  reduction.
\newblock \emph{arXiv preprint arXiv:1802.03426}.

\bibitem[{Petegrosso, Li, and Kuang(2020)}]{petegrosso2020machine}
Petegrosso, R.; Li, Z.; and Kuang, R. 2020.
\newblock Machine learning and statistical methods for clustering single-cell
  RNA-sequencing data.
\newblock \emph{Briefings in bioinformatics}, 21(4): 1209--1223.

\bibitem[{Ren et~al.(2020)Ren, Wang, Li, and Xu}]{ren2020deep}
Ren, Y.; Wang, N.; Li, M.; and Xu, Z. 2020.
\newblock Deep density-based image clustering.
\newblock \emph{Knowledge-Based Systems}, 105841.

\bibitem[{Rodriguez and Laio(2014)}]{denpeak}
Rodriguez, A.; and Laio, A. 2014.
\newblock Clustering by fast search and find of density peaks.
\newblock \emph{Science}, 344(6191): 1492--1496.

\bibitem[{Shi and Malik(2000)}]{ncut}
Shi, J.; and Malik, J. 2000.
\newblock Normalized cuts and image segmentation.
\newblock \emph{TPAMI}, 22(8): 888--905.

\bibitem[{Simonyan and Zisserman(2015)}]{Simonyan15}
Simonyan, K.; and Zisserman, A. 2015.
\newblock Very Deep Convolutional Networks for Large-Scale Image Recognition.
\newblock In \emph{ICLR}.

\bibitem[{Tsai, Li, and Zhu(2020)}]{tsai2020mice}
Tsai, T.~W.; Li, C.; and Zhu, J. 2020.
\newblock MiCE: Mixture of Contrastive Experts for Unsupervised Image
  Clustering.
\newblock In \emph{ICLR}.

\bibitem[{Van~Gansbeke et~al.(2020)Van~Gansbeke, Vandenhende, Georgoulis,
  Proesmans, and Van~Gool}]{van2020scan}
Van~Gansbeke, W.; Vandenhende, S.; Georgoulis, S.; Proesmans, M.; and Van~Gool,
  L. 2020.
\newblock Scan: Learning to classify images without labels.
\newblock In \emph{ECCV}, 268--285. Springer.

\bibitem[{Xie, Girshick, and Farhadi(2016)}]{xie2016unsupervised}
Xie, J.; Girshick, R.; and Farhadi, A. 2016.
\newblock Unsupervised deep embedding for clustering analysis.
\newblock In \emph{ICML}, 478--487.

\bibitem[{Xu, Liu, and Gong(2003)}]{xu2003document}
Xu, W.; Liu, X.; and Gong, Y. 2003.
\newblock Document clustering based on non-negative matrix factorization.
\newblock In \emph{International ACM SIGIR conference on Research and
  development in information retrieval}, 267--273.

\bibitem[{Yang et~al.(2017)Yang, Fu, Sidiropoulos, and Hong}]{dcn}
Yang, B.; Fu, X.; Sidiropoulos, N.~D.; and Hong, M. 2017.
\newblock Towards k-means-friendly spaces: Simultaneous deep learning and
  clustering.
\newblock In \emph{ICML}, 3861--3870. PMLR.

\bibitem[{Yang, Parikh, and Batra(2016)}]{jule}
Yang, J.; Parikh, D.; and Batra, D. 2016.
\newblock Joint unsupervised learning of deep representations and image
  clusters.
\newblock In \emph{CVPR}, 5147--5156.

\bibitem[{Yang et~al.(2019)Yang, Deng, Zheng, Yan, and Liu}]{yang2019deep}
Yang, X.; Deng, C.; Zheng, F.; Yan, J.; and Liu, W. 2019.
\newblock Deep spectral clustering using dual autoencoder network.
\newblock In \emph{CVPR}, 4066--4075.

\bibitem[{Yang et~al.(2010)Yang, Xu, Nie, Yan, and Zhuang}]{ldmgi}
Yang, Y.; Xu, D.; Nie, F.; Yan, S.; and Zhuang, Y. 2010.
\newblock Image clustering using local discriminant models and global
  integration.
\newblock \emph{TIP}, 19(10): 2761--2773.

\bibitem[{Yaohui, Zhengming, and Fang(2017)}]{yaohui2017adaptive}
Yaohui, L.; Zhengming, M.; and Fang, Y. 2017.
\newblock Adaptive density peak clustering based on K-nearest neighbors with
  aggregating strategy.
\newblock \emph{Knowledge-Based Systems}, 133: 208--220.

\bibitem[{Zelnik-Manor and Perona(2005)}]{stsc}
Zelnik-Manor, L.; and Perona, P. 2005.
\newblock Self-tuning spectral clustering.
\newblock In \emph{NeurIPS}, 1601--1608.

\bibitem[{Zou et~al.(2020)Zou, Lin, Jiang, Liu, and Zeng}]{zou2020sequence}
Zou, Q.; Lin, G.; Jiang, X.; Liu, X.; and Zeng, X. 2020.
\newblock Sequence clustering in bioinformatics: an empirical study.
\newblock \emph{Briefings in bioinformatics}, 21(1): 1--10.

\end{thebibliography}
}

\newpage
\section{Supplementary Material}

\section{Proof of Lemma 1}

\begin{proof}
Let $\mathcal{X}_{S}$ be a set of samples in the intersection between separable subsets $\mathcal{X}_{I}$ and $\mathcal{X}_{J}$ with low density, and $\mathbf{X}_{s}$ is an interior point of $\mathcal{X}_{S}$, and assume that data samples in those three subsets are uniformly distributed with densities $\theta_{I}$, $\theta_{J}$ and $\theta_{S}$, (i.e., $\theta_{I} > \theta_{S}$ and $\theta_{J} > \theta_{S}$). 
Let $\mathcal{D}_{S}$ be the set of top $M$ nearest distances around $\mathbf{X}_{s}$ and  $r_{RI}$, $r_{RJ}$, and $r_{RS}$ be the radius of regions covered to define top $M$ edges for $\mathbf{X}_{i}$, $\mathbf{X}_{j}$ and $\mathbf{X}_{s}$. Since $\theta_{I} > \theta_{S}$ and $\theta_{J} > \theta_{S}$, we get $\mu_{SI} = \frac{r_{RS}}{r_{RI}} > 1$ and $\mu_{SJ}=\frac{r_{RS}}{r_{RJ}} > 1$. According to Definition~\ref{definition1}, $\mathcal{X}_{R}$ has two subsets: a subset from $\mathcal{X}_{I}$ or $\mathcal{X}_{J}$ and a subset from $\mathcal{X}_{S}$, i.e., $\mathcal{X}_{R} \cap \mathcal{X}_{I} \neq \emptyset$ or $\mathcal{X}_{R} \cap \mathcal{X}_{J} \neq \emptyset$. 

If $\mathcal{X}_{R} \cap \mathcal{X}_{I} \neq \emptyset$, 
then the variance of $\mathcal{D}_{T}$ is given as
\begin{eqnarray}
\small
\label{eq:lemma1}
    Var(\mathcal{D}_{T}) = \frac{1}{M} \bigg(n_{0}Var(\mathcal{D}_{I})  + (M-n_{0})Var(\mathcal{D}_{S}) + \nonumber\\   
      n_{0} (\bar{\mathcal{D}_{I}} - \bar{\mathcal{D}_{T}})^{2} + 
     (M-n_{0})(\bar{\mathcal{D}_{S}} - \bar{\mathcal{D}_{T}})^{2} \bigg)
\end{eqnarray}
where $\bar{\mathcal{D}_{I}}$, $\bar{\mathcal{D}_{S}}$ and $\bar{\mathcal{D}_{T}}$ represent the mean distances of $\mathcal{D}_{I}$, $\mathcal{D}_{S}$ and $\mathcal{D}_{T}$, and $n_{0}$ is the number of points in $\mathcal{X}_{R} \cap \mathcal{X}_{I}$. 
As we have assumed that the samples in $\mathcal{X}_{I}$ and $\mathcal{X}_{S}$ are uniformly distributed, $\mathcal{D}_{S} = \mu_{SI} \mathcal{D}_{I}$ which leads to $Var(\mathcal{D}_{S}) = \mu_{SI}^{2} Var(\mathcal{D}_{I}) > Var(\mathcal{D}_{I})$. 
Therefore, \eqref{eq:lemma1} is rewritten as 
\begin{eqnarray}
\small
\label{eq:lemma1-2}
    Var(\mathcal{D}_{T}) \ge  \frac{n_{0}Var(\mathcal{D}_{I})  + (M-n_{0})Var(\mathcal{D}_{S})}{M}  > \nonumber\\
    \frac{n_{0}Var(\mathcal{D}_{I})  + (M-n_{0})Var(\mathcal{D}_{I})}{M} = Var(\mathcal{D}_{I})
\end{eqnarray}
Similarly, if $\mathcal{X}_{R} \cap \mathcal{X}_{J} \neq \emptyset$, we can obtain $Var(\mathcal{D}_{T}) > Var(\mathcal{D}_{J})$. 
\end{proof}

\section{Details of Experiments on Challenging Image Datasets}

\paragraph{Image Datasets} 
In the main manuscript, we compared our VCC with other recent deep clustering algorithms on three challenging image datasets: CIFAR10 \cite{krizhevsky2009learning}, CIFAR100-20 \cite{krizhevsky2009learning} and STL10 \cite{coates2011analysis}. The summary of these three datasets is shown in TABLE~\ref{tab:image_datasets}.

\begin{table}[!h]
\centering
\caption{\footnotesize Summary of Image Datasets. }
\label{tab:image_datasets}
\scalebox{.7}{
\begin{tabular}{ l|ccc }
\toprule
Dataset & \# Samples (train; val) & \# Classes & Image Size ($H\times W \times C$) \\ \cmidrule{1-4}
CIFAR10 & (50, 000; 10, 000) & 10 & $32\times 32 \times 3$ \\ 
CIFAR100-20 & (50, 000; 10, 000) & 20 & $32\times 32 \times 3$ \\
STL10  & (5,000; 8, 000) & 10 & $96\times 96 \times 3$ \\ 
\bottomrule
\end{tabular}}
\vspace{-10pt}
\end{table}

\paragraph{Experiment Settings.} For fair comparisons, the experiment settings follow the most current state-of-the-art deep clustering method SCAN~\cite{van2020scan}. The model is first trained with the training dataset and then evaluated with the validation dataset. When comparing SCAN and VCC, they both use the same embedding with dimension 128 from pretrained SimCLR with ResNet18 as the backbone. The parameters used in VCC have been described in the ``Implementation Details" section in the main manuscript.

\section{Ablation Study: Performance and Losses}

\begin{figure}[!t]
  \centering
      \begin{subfigure}{.48\textwidth}
        \centering
            \includegraphics[width=.8\linewidth]{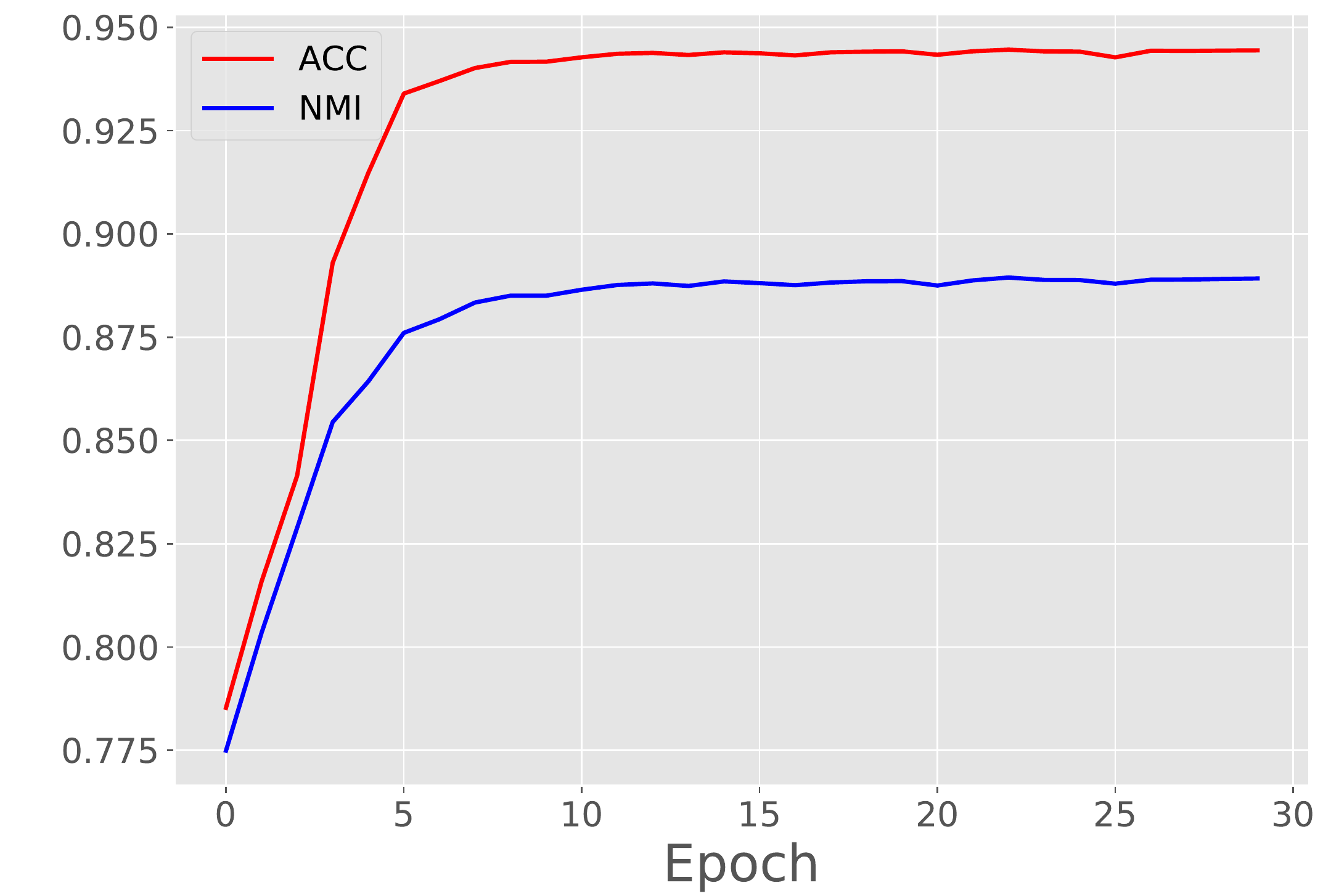}
            \vspace{-5pt}
        \caption{\footnotesize Performance measures.}
    \end{subfigure}
    \begin{subfigure}{.48\textwidth}
        \centering
            \includegraphics[width=.8\linewidth]{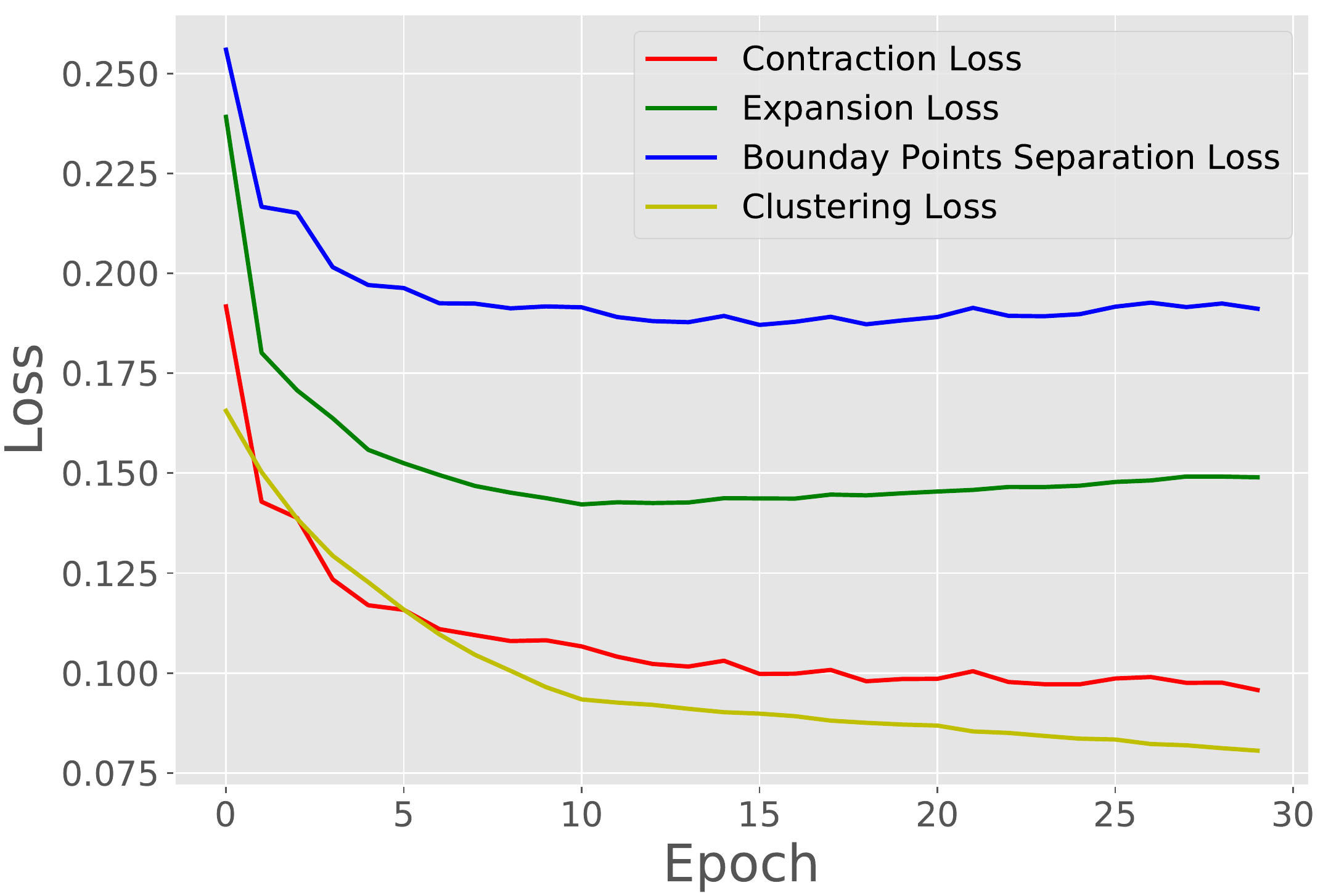}
            \vspace{-5pt}
        \caption{\footnotesize Loss/Regularizers.}
    \end{subfigure}
    
    \caption{\footnotesize Training process of VCC w.r.t. epoch on MNIST-test data. Top: Changes of accuracy and NMI, Bottom: Change of losses. 
    The training is stable and losses/regularizers converge fast in the early stages. 
    }
    \label{fig:acc_loss}
    \vspace{-15pt}
\end{figure}

Fig.~\ref{fig:acc_loss} (a) plots changes of ACC and NMI during the training on MNIST-test w/o VGG, and
(b) shows the changes of individual losses and regularizers introduced in the method section. 
We can see that loss and regularizers converge fast robustly. 
Even after the accuracy and NMI reach their peaks, clustering loss continues to decrease to make the final clusters more localized.

\section{Ablation Study: Effects of the Dimension of Latent Space}

\begin{table}[!h]
\centering
\caption{\footnotesize VCC clustering performance comparison with different dimensions of latent space on MNIST-test dataset. }
\label{tab:supp_latent_dim}
\scalebox{.62}{
\begin{tabular}{ l|ll|ll|ll|ll|ll }
\toprule
& \multicolumn{2}{c}{dim=2} & \multicolumn{2}{c}{dim=10} & \multicolumn{2}{c}{dim=20} & \multicolumn{2}{c}{dim=50} & \multicolumn{2}{c}{dim=100}  \\ \cmidrule{2-11}
&  ACC & NMI & ACC & NMI & ACC & NMI & ACC & NMI & ACC & NMI \\ \midrule
VCC  & 0.946  &  0.889 &  0.938  &  0.884 &  0.947  & 0.892  &  0.944  &  0.888 &  0.945  &  0.889   \\ 
\bottomrule
\end{tabular}}
\end{table}

In the main manuscript, the dimension of the latent space is set to 2 for visualization purposes. In this ablation study, we change the dimension of the latent space from 2 to 100 to evaluate VCC's clustering performance. TABLE~\ref{tab:supp_latent_dim} shows that VCC's performances change little with different dimensional latent space. That is, with our method, a simple MLP with a low-dimensional latent space is enough to get good and stable clustering results.

\end{document}